\documentclass[10pt,twocolumn,letterpaper]{article}

\usepackage{wacv}              

\newcommand{\chinese}[1]{\begin{CJK}{UTF8}{gbsn}#1\end{CJK}}

\usepackage{booktabs}
\usepackage{CJKutf8}
\usepackage{multirow}

\definecolor{wacvblue}{rgb}{0.21,0.49,0.74}
\usepackage[pagebackref,breaklinks,colorlinks,allcolors=wacvblue]{hyperref}
\usepackage{placeins}

\def\wacvPaperID{2877} 
\def\confName{WACV}
\def\confYear{2027}

\title{Using OCR Heads to Verbalize Image Semantics}

\author{Sheridan Feucht \\
Northeastern University 
\and
Benno Krojer \\
Northeastern University \\
\and
Sarah Wang \\
Independent
\and
Henry Abrahamsen \\ 
Northeastern University
\and 
Byron C. Wallace \\ 
Northeastern University 
\and
David Bau \\
Northeastern University
}

\begin{document}
\maketitle
\begin{abstract}
How do VLMs map from pixels to semantics? To understand this general question, we focus on a narrow one: studying how VLMs perform optical character recognition (OCR). Across four models, we identify attention heads causally necessary for OCR, and discover that these are in fact general-purpose heads that output interpretable semantic features across all image tokens. For example, pointing these heads at an image token containing the word ``bike'' causes Qwen3-VL-8B to output ``bike,'' but pointing them at a bird wing causes the model to output the token ``feathers.'' We collapse these heads' attention weights into a single \textbf{verbalization lens} transformation that reveals interpretable semantic features in hidden states across all layers. When combined with projection to vocabulary space, we can obtain interpretable labels starting from layer 0, showing that image representations are in fact aligned with language in early layers. We find that we can also use the inverse of this transformation to edit non-word concepts, e.g., replacing a tractor with a revolver in a naturalistic image, providing causal evidence that this subspace is useful for more than just OCR. Our results are an example of how the study of specific mechanisms can shed light on broader interpretability problems.\footnote{
For code and interactive demos, see \href{https://ocr.baulab.info}{https://ocr.baulab.info}. 
Correspondence: {\tt\small feucht.s@northeastern.edu}}
\end{abstract}

\section{Introduction}\label{sec:intro}

\begin{figure}
    \centering
    \includegraphics[width=\linewidth]{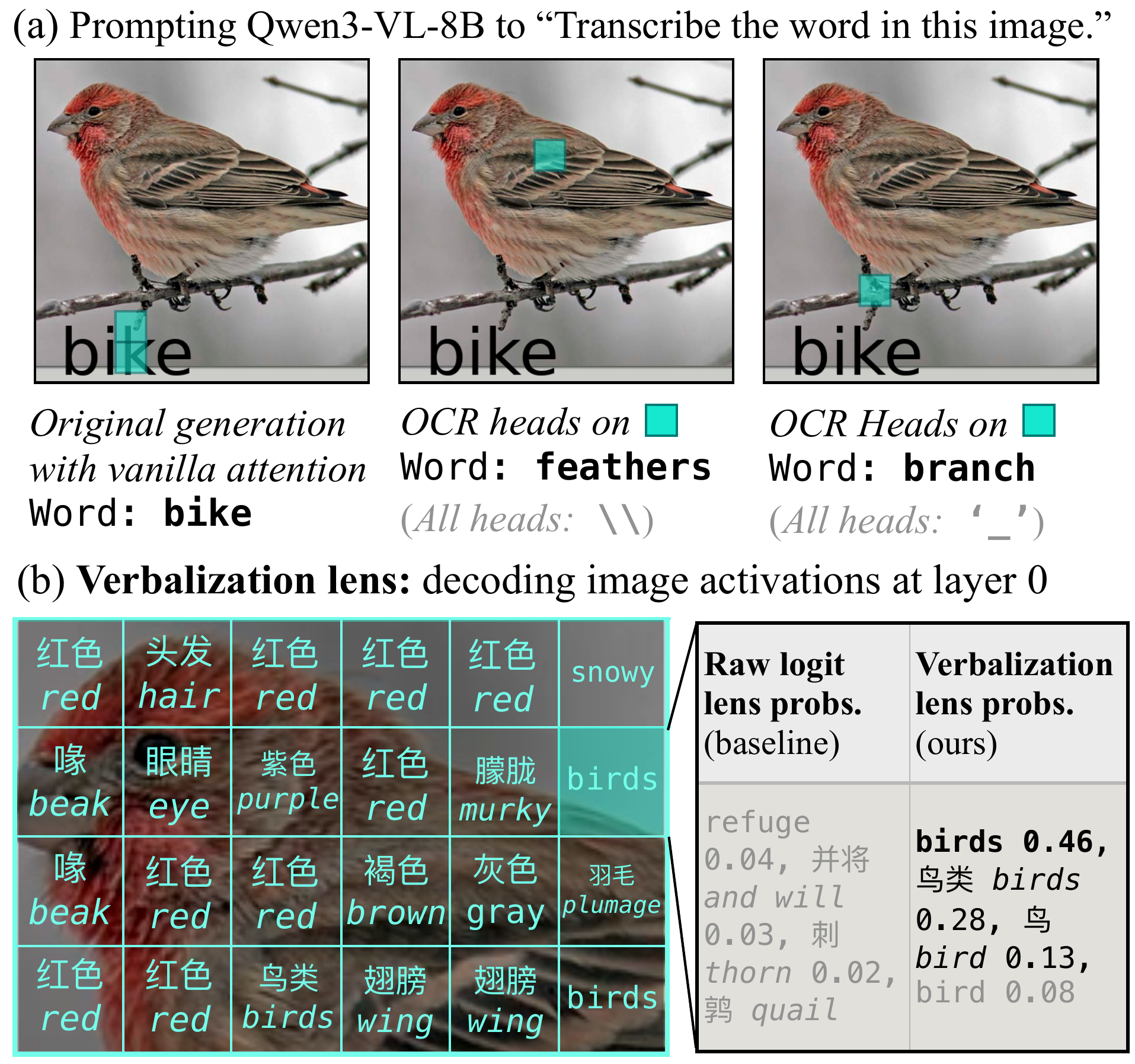}
    \caption{We find that VLM attention heads responsible for OCR are actually generic \textit{verbalization heads} that can read out semantics of image tokens that do \textit{not} contain text. (a) Verbalization heads attend to text when prompted to transcribe words, but manually redirecting their attention to image tokens that do \textit{not} contain text causes the model to output words corresponding to the semantics of those tokens (top 10\% of Qwen3-VL-8B verbalization heads). (b) We repurpose these heads' attention weights to create a \textbf{verbalization lens} that can be applied to image representations from any layer. Here, our approach reveals linguistic information linearly encoded in image tokens starting from layer 0, which logit lens is unable to see.}
    \label{fig:figure1}
\end{figure}

Reading is a specialized form of perception that involves mapping from very specific visual forms to semantic concepts. 
Because it is closely tied to language, it occupies a special place in neuroscience research, with a large body of work studying specific areas of the brain used to represent word forms \cite{McCandliss2003, DEHAENE2011254, dehaenereview}. The central point of interest is the interface between \emph{visual} and \emph{linguistic} representations: a question that is also a concern for AI interpretability \cite{venhoff2025visualrepresentationsmaplanguage, neo, schwettmann2023}. 

In this work we study how vision-language models ``read'' (i.e., perform optical character recognition, or OCR) \cite{karamolegkou2026readingguessingvisualgrounding,
liang2026visualmeritlinguisticcrutch,
he2025seeingbelievingmitigatingocr}, which we show can help us understand how these models map from pixels to semantics in general. 
We discover that attention heads responsible for OCR can also be used to verbalize semantic information across a wide range of image tokens. We first isolate a set of attention heads that are causally responsible for OCR; if we prompt Qwen3-VL-8B \citep{bai2025qwen3vltechnicalreport} to transcribe a word in, e.g., Figure~\ref{fig:figure1}a, it will correctly output ``bike,'' and we can observe these heads attending to the tokens that correspond to that word. However, if we intervene on these attention heads to attend to a bird wing, we find that Qwen3-VL-8B will output the token \texttt{\_feathers}. Similarly, it will output \texttt{\_branch} if we force these heads to attend to a branch. Due to these heads' ability to describe image tokens that do not contain text, we refer to them as general \emph{verbalization heads}.

We study the subspace of activation space read by verbalization heads across four models, and find that it contains a wealth of high-level semantic information about inputted images. Specifically, we combine the parameters of identified verbalization heads into a single matrix that can be applied to any hidden state, following techniques described in prior work on language-only models \cite{elhage2021mathematical, feucht2025dualroute, feucht2025arithmetic}. When combined with logit lens \cite{nostalgebraist, neo, jiang} (which projects hidden activations to vocabulary space), our verbalization matrix reveals semantic information across all image tokens, as seen in Figure~\ref{fig:figure1}b.

Unlike vanilla logit lens, our approach reveals information about image tokens starting from layer 0 of the language backbone. This means that when images are passed from the image encoder to the language decoder, they already linearly encode linguistic information. In other words, our results provide evidence that the image encoder immediately supplies vision representations with linguistic information \cite{latentlens, featurevizaligned}, despite the apparent modality gap in early layers claimed in prior work \cite{jiang2024hallucinationaugmentedcontrastivelearning, schwettmann2023, venhoff2025toolatetorecall, venhoff2025visualrepresentationsmaplanguage, NEURIPS2024_ec3c79dc}.

These verbalizations are not just correlational, but can also be used to \emph{edit} latent image representations for Qwen3-VL models. By adding and subtracting latent vectors within the subspace read by verbalization heads, we can edit a model's activations to replace objects in an image, e.g., causing Qwen3-VL-2B to describe an ant as ``an iPod resting on a green plant stem'' (Figure~\ref{fig:edit_main}a). This indicates that the verbalizations obtained using our approach are causally important for more than just OCR. 

Although reading letters appears to be a narrow task, it provides a clean way to identify more general VLM components that map from vision to language space. Obtaining supervision for OCR tasks is uniquely straightforward: while an image of a bird might be mapped to the concepts ``animal,'' ``feathers,'' or ``small,'' an image of the letters b-i-r-d is uniquely associated with the word ``bird.'' Thus, identifying attention heads in this tractable setting allows us to understand the semantic representations of VLMs more broadly. We provide code and an interactive demo of verbalization lens across models at \href{https://ocr.baulab.info}{https://ocr.baulab.info}.

\section{Finding Verbalization Heads}\label{sec:finding-heads}

\subsection{Approach}

\begin{figure}
    \centering
    \includegraphics[width=\linewidth]{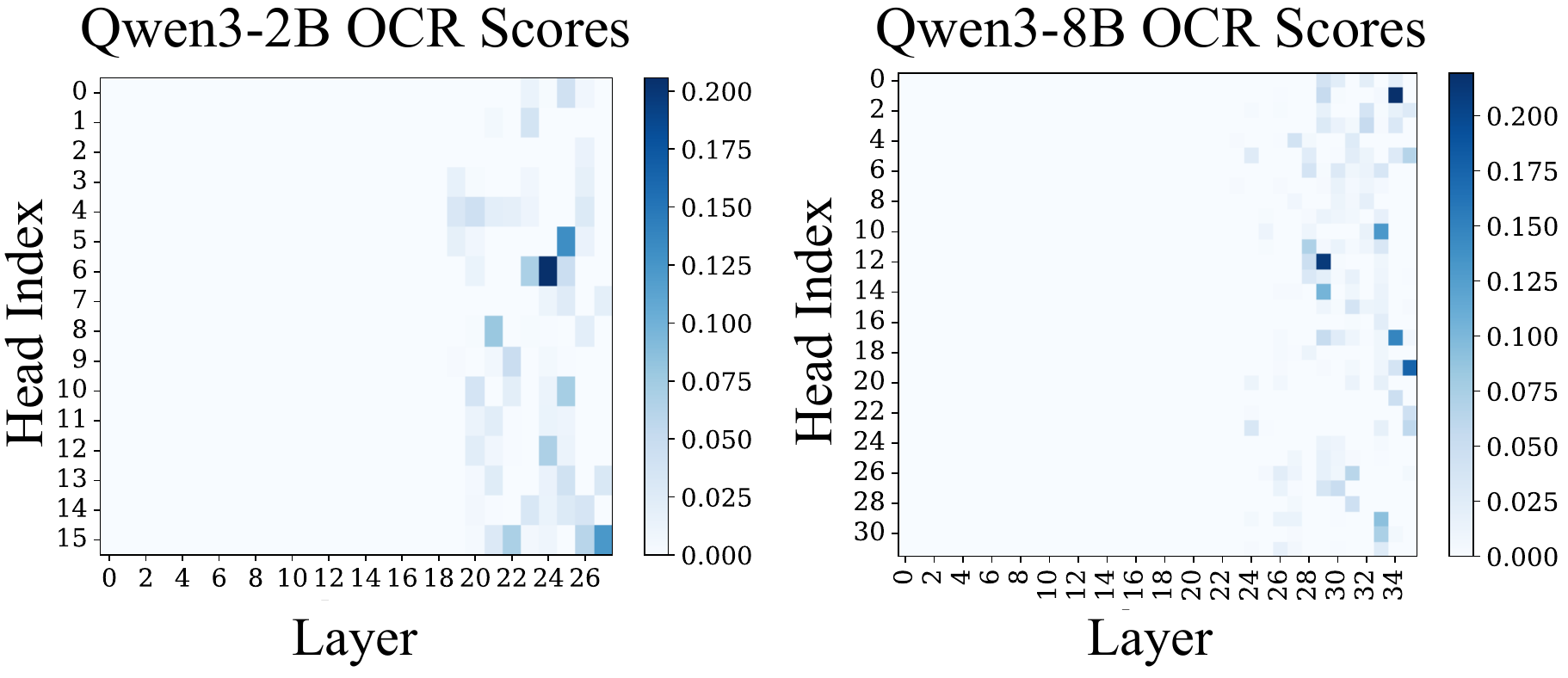}
    \caption{OCR Scores (Equation~\ref{eq:ocr_score}) for all heads in Qwen3-VL models. We prompt models to transcribe random words $w_i$ in images $x_i$ and measure $P(w_i)$ according to logit lens at the final token position for each individual head ($n$=1024). Many heads in late layers promote the word contained in an OCR dataset image. See Figure~\ref{fig:additional-heatmaps} for scores for Molmo2-7B and Llava-Next-34B.}
    \label{fig:qwen-heatmaps}
\end{figure}

\begin{figure}
    \centering
    \includegraphics[width=\linewidth]{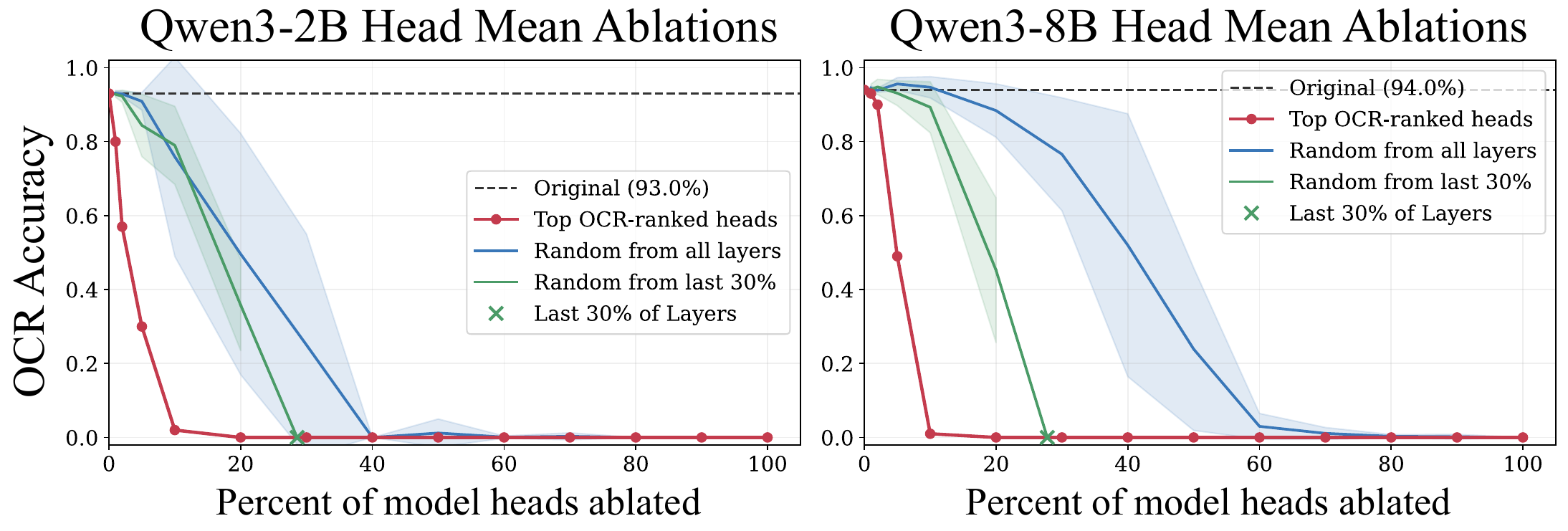}
    \caption{Top-scoring attention heads from Figure~\ref{fig:qwen-heatmaps} are causally necessary for doing OCR. Across models, we find that heads responsible for OCR comprise 10\% of total attention heads. These ablations are run for 100 images across 10 seeds for random baselines; see Figure~\ref{fig:other-model-ablations} for Molmo2-7B and Llava-Next-34B.}
    \label{fig:head-ablations}
\end{figure}

We first search for attention heads causally responsible for extracting text from image representations. 
We set up a simple OCR task by placing random English words on top of ImageNet \cite{imagenet} images.\footnote{We use image backgrounds because OCR performance is poor for words on blank backgrounds; see Table~\ref{tab:ocr_acc} in Appendix.} 
We filter words, keeping only those corresponding to a single token in the model vocabulary, then pre-fill the input with ``Word:'' and read out the response. Table~\ref{tab:ocr_acc} shows that all models we study can reliably complete our OCR task; see Figure~\ref{fig:ocr_dataset_examples} for dataset examples. 

To find attention heads that are responsible for OCR, we apply logit lens \cite{nostalgebraist, neo, jiang} to each head's output and measure the probability of the desired word. 
Each image in our OCR dataset $x_i\in\mathcal{X}_{\text{OCR}}$ contains a word $w_i$.
Let $\mathbf{h}_{x_i}^{(l,h)}\in\mathbb{R}^{d_{\text{head}}}$ refer to the output of attention head $h$ at layer $l$ at the final token position for image $x_i$, when the model is about to output its transcription. We project this vector to the residual stream dimension using the output projection for that head $\mathbf{W}_O^{(l,h)}\in\mathbb{R}^{d_{\text{model}}\times d_\text{head}}$, and directly decode to vocabulary space to obtain $P(w_i)$. We take the mean of this value across images:

\begin{equation}\label{eq:ocr_score}
    s(l,h)=\frac{1}{|\mathcal{X}_{\text{OCR}}|}\sum_{x_i\in\mathcal{X}_{\text{OCR}}}\textsc{LogitLens}(\mathbf{W}_O^{(l,h)}\mathbf{h}_{x_i}^{(l,h)})[w_i]
\end{equation}

\noindent which yields a score $s(l, h)$ for attention head $h$ at layer $l$, measuring how strongly this head outputs words $w_i$ across the OCR dataset. Logit lens \cite{nostalgebraist} is defined as: 

\begin{equation}\label{eq:ll}
    \textsc{LogitLens}(\mathbf{a}) = \text{softmax}(\mathbf{W}_U\text{norm}(\mathbf{a}))
\end{equation}

\noindent for a hidden state $\mathbf{a}\in\mathbb{R}^{d_{\text{model}}}$, where ``norm'' refers to the final norm, and $\mathbf{W}_U \in\mathbb{R}^{{|\mathcal{V}|}\times d_{\text{model}}}$ is the ``unembedding'' matrix mapping to the token vocabulary space  $\mathcal{V}$. 
The softmax induces a distribution over tokens. 

The intuition behind our approach is that when the model is about to transcribe a word, some set of attention heads are responsible for ``fetching'' $w_i$ from the image and bringing that information into the column space of the final decoder head, similar to OCR heads found in prior work \citep{baek}.
Equation~\ref{eq:ocr_score} is designed to find these heads, assuming they exist.

\subsection{Results}
\label{sec22}

Figure~\ref{fig:qwen-heatmaps} shows OCR scores for all attention heads in two Qwen3-VL models. Notably, heads with high scores are concentrated in late layers. 
To assess whether these heads are causally important for OCR, we progressively mean-ablate\footnote{To ``mean-ablate'' \cite{ioi, feucht2025dualroute} an attention head, we replace the output of that head across all token positions with its mean over 1000 examples from our OCR dataset (also taken across token positions).} heads with the highest scores and measure the trend in task performance in Figure~\ref{fig:head-ablations}. Across all models, mean-ablating the top 10\% of heads ranked by OCR scores fully degrades task performance. OCR heads are concentrated in late layers, which might be a confound, but ablating an equivalent number of random late-layer heads is less effective. This suggests that OCR heads are most responsible for the model's performance on this task. 
Results are similar for Molmo2-7B and Llava-Next-34B (see Appendix~\ref{app:score-other-models}). 

\subsection{Head Behavior on Non-Text}

\begin{figure}
    \centering
    \includegraphics[width=\linewidth]{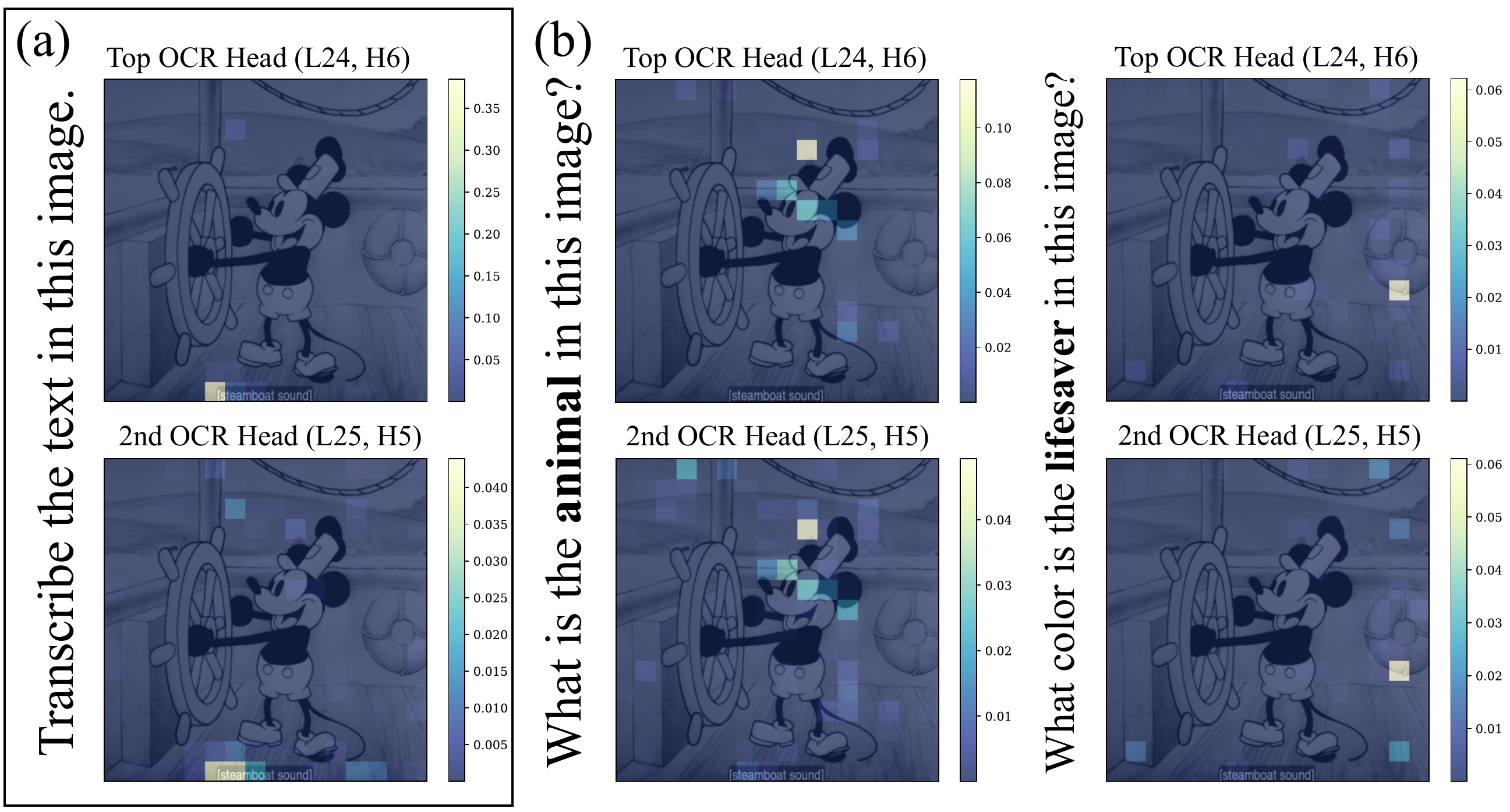}
    \caption{Top-scoring heads for OCR attend to text---but under different prompt settings, they can also attend to animals and objects. We show the two top-OCR-scoring attention heads in Qwen3-VL-2B as an example. (a) These heads attend to the subtitle ``[steamboat sound]'' when prompted to transcribe text. (b) The same heads also attend to a mouse when prompted ``What is the animal in this image?'' and a lifesaver when prompted ``What color is the lifesaver in this image?'' (Qwen3-VL-2B's answers to these questions are ``Mouse'' and ``White,'' respectively). This behavior suggests that these heads are not purely dedicated to OCR.}
    \label{fig:switching-attn}
\end{figure}

Although we identify these heads based on a small synthetic OCR task, they appear to be broader \emph{verbalization heads} used for more general image understanding (see Figure~\ref{fig:figure1}). 
While delineating the exact purpose of these heads is beyond the scope of this work, we now briefly discuss some intuition for why these heads are able to verbalize non-textual tokens. 

In Figure~\ref{fig:switching-attn}a, we visualize attention patterns for top-scoring Qwen3-VL-2B OCR heads on a random image. 
As expected, these heads attend to image tokens containing text when prompted to transcribe the text in the image. However, we also find that when the model is prompted ``What is the animal in this image?,'' the same attention heads begin to attend to tokens corresponding to the mouse in the image, rather than the subtitles. 
This attention behavior is analogous to work on \emph{filter heads} in LLMs \cite{sensharma2023filter, andrewleeqk2026}, which ``retrieve'' items from the prior context based on a prompt query (see Appendix~\ref{app:overlap} for discussion).

What do verbalization heads output when they attend to non-text tokens? 
Figure~\ref{fig:figure1}a shows a qualitative example for Qwen3-VL-8B: redirecting the top 10\% of verbalization heads to attend to, e.g., a token containing a bird wing causes the model to output the token \texttt{\_feathers}. This motivates a question: can we use these attention heads to create a ``verbalizer'' for image tokens in general?

We note that it is likely one could also identify this set of attention heads using tasks other than OCR, e.g., asking ``What is in this image?'' with supervision from image annotations. 
However, OCR provides a straightforward one-to-one mapping from visual inputs to language outputs. 
If we were to run the same procedure with natural images, a given image token of a bird might plausibly map to many different tokens, including ``bird,'' ``feathers,'' ``cardinal,'' ``wing,'' or ``\chinese{小鸟}'' (``bird'' in Chinese). The OCR task offers unambiguous supervision, enabling broader understanding of how VLMs map from pixels to semantics. 

\section{Building a Verbalization Lens}

\begin{figure*}
    \centering
    \includegraphics[width=\linewidth]{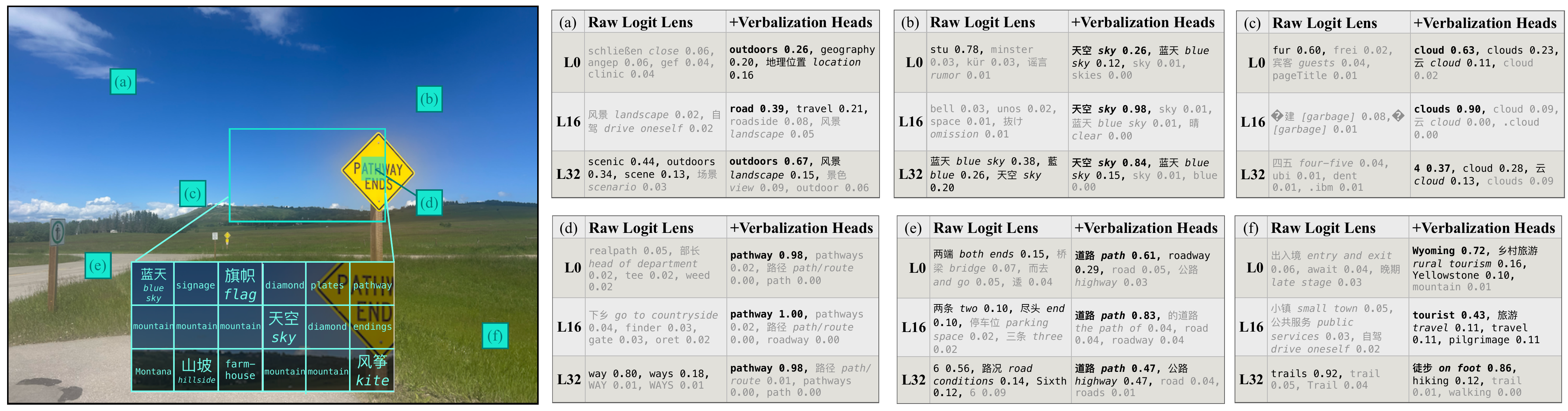}
    \caption{Verbalization lens reveals semantic information contained within VLM image tokens starting from early layers. We show several token readouts for one image for Qwen3-VL-8B, comparing raw logit lens to the top 10\% of verbalization heads. Applied to image tokens that contain text, verbalization lens shows tokens corresponding to the appropriate word; see subfigure (d). Applied to other image tokens, verbalization lens reveals interpretable semantic information, outputting \texttt{\_cloud} for a token containing clouds (c), or more general information like \texttt{\_outdoors} above a possible register token \cite{darcet2024register} in (a). Outputs are often in Chinese for Qwen3-VL models, likely because they are Chinese models. Gray text means probability is below 0.10; top verbalization lens predictions are bolded for emphasis. See Appendix~\ref{app:lens_qual_allmodels} for qualitative results across models.}
    \label{fig:lens_qual}
\end{figure*}

\subsection{Verbalization Transformation}

Instead of redirecting attention maps as in Figure~\ref{fig:figure1}a (which requires a separate forward pass for every image token), we compact the parameters of these verbalization heads into a single linear transformation that can reveal semantic information contained within any hidden state. Specifically, we sum the output-value (OV) matrices \cite{elhage2021mathematical} of verbalization heads across layers.
This is roughly equivalent to intervening on attention maps, except that heads can ``attend'' to activations from any layer \cite{feucht2025dualroute, feucht2025arithmetic}.

Let $\mathcal{O}_p$ be a set containing the top-$p$\% of OCR heads; across models, we use $p$=10\% based on Sec. \ref{sec22}. We construct a matrix $\mathbf{L}_p\in\mathbb{R}^{d_{\text{model}}\times d_{\text{model}}}$ comprising the parameters of these verbalization heads. We calculate

\begin{equation}\label{eq:build_ocr}
    \mathbf{L}_p = \sum_{(l,h)\in\mathcal{O}_p} \mathbf{W}_O^{(l,h)}\mathbf{W}_V^{(l,h)},
\end{equation}

\noindent based on the output and value projections of heads in $\mathcal{O}_p$, which have shape $\mathbf{W}_O^{(l,h)}\in\mathbb{R}^{d_{\text{model}}\times d_{\text{head}}},\mathbf{W}_V^{(l,h)}\in\mathbb{R}^{d_{\text{head}}\times d_{\text{model}}}.$ 
Multiplied together, each ``OV matrix'' determines what this head will output if it attends to a particular token; adding them together is equivalent to summing the outputs of multiple heads. Thus, applying this matrix to a hidden state is equivalent to ``forcing'' all heads in $\mathcal{O}_p$ to attend to that hidden state and summing the resulting outputs. 

\subsection{Decoding to Vocabulary Space}

After applying the verbalization transformation from Equation~\ref{eq:build_ocr} to an activation $\mathbf{a}^{(l)}\in\mathbb{R}^{d_{\text{model}}}$, we can project the resulting vector to vocabulary space (Equation~\ref{eq:ll}) to obtain probabilities for each token:

\begin{equation}\label{eq:ocr_lens}
    \textsc{VerbalLens}(\mathbf{a}^{(l)}) = \textsc{LogitLens}(\mathbf{L}_p\mathbf{a}^{(l)}).
\end{equation}

This approach can be thought of as a quick approximation of the proof-of-concept in Figure~\ref{fig:figure1}a. There are two key differences: first, heads across layers can ``attend'' to $\mathbf{a}^{(l)}$ regardless of whether they are also at layer $l$. This still yields interpretable results, perhaps because layer order has been shown to be mostly interchangeable in LLMs \cite{ladstages2025, todd2024function, feucht2025dualroute}.
Second, we directly decode to vocabulary space, rather than allowing the model to further process these head outputs. This gives us the advantage of being able to run a single forward pass and decode any activation with two simple matrix multiplications (the naive approach requires a separate forward pass for each image token). We can conceptualize $\mathbf{L}_p$ as an operation that ``focuses'' logit lens on the correct subspace of model activations.

\begin{figure*}
    \centering
    \includegraphics[width=\linewidth]{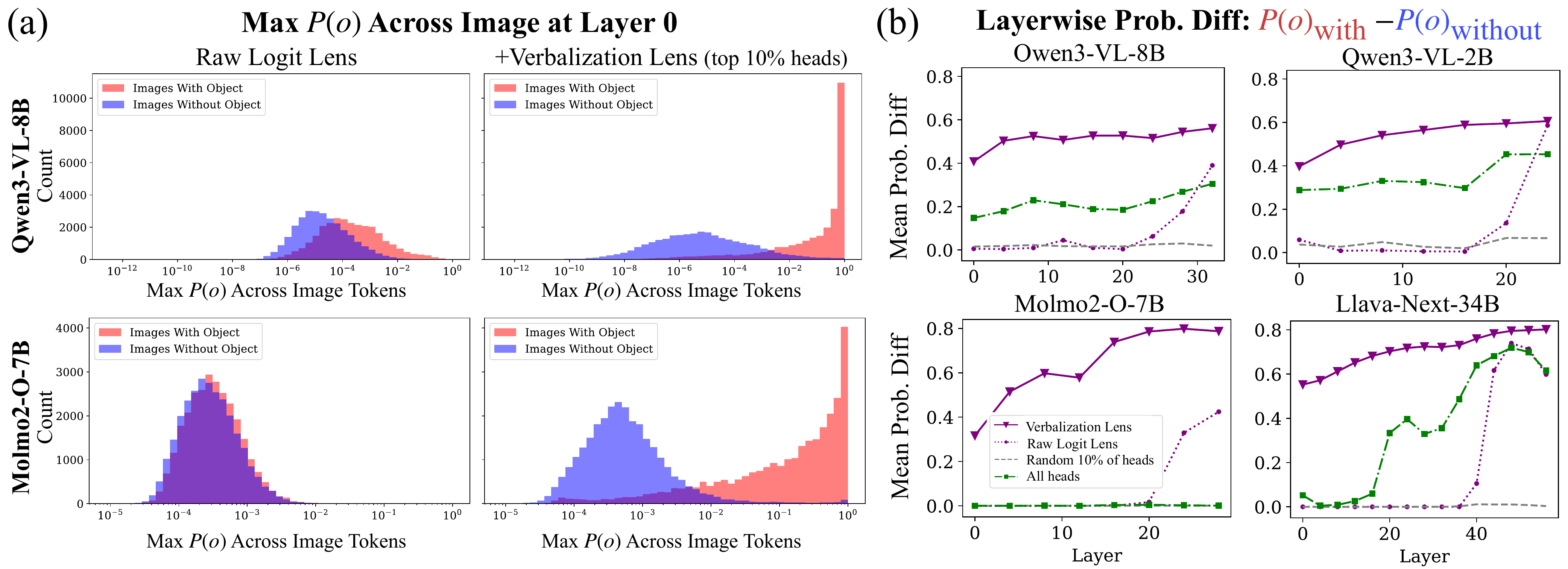}
    \caption{Verbalization lens is sensitive to whether an image contains an object: it yields high probabilities for images that contain objects, and low probabilities for images that do not. We apply logit lens (Equation~\ref{eq:ll}) to VLM representations of images from the COCO dataset \cite{COCO} with and without this transformation, and extract the probability of the COCO class label token for all single-token categories, similar to \cite{jiang}. (a) At layer 0, verbalization lens is better able to detect whether an image contains an object---$P(o)$ is consistently higher for the red histogram for Qwen3-VL-8B and Molmo2-7B. (b) Average difference in maximum $P(o)$ across layers for images containing an object vs. random images not containing that object. Verbalization lens increases the gap between the two settings more than raw logit lens, especially in early-middle layers. Using the weights of \textit{all} attention heads gives a weaker effect than just verbalization heads.
    }
    \label{fig:detection}
\end{figure*}

\subsection{Qualitative Results}
Figure~\ref{fig:lens_qual} shows an example of logit lens applied to image hidden states across layers \emph{with} and \emph{without} our verbalization transformation. Applying our verbalization heads reveals semantic information both for image tokens containing written text (e.g., \texttt{\_pathway} for a patch containing the text ``PATHWAY'', Figure~\ref{fig:lens_qual}d) and image tokens containing more generic visual information (e.g., \texttt{\_cloud} for a patch containing clouds, Figure~\ref{fig:lens_qual}c). 
Our approach yields interpretable results as soon as layer 0, whereas raw logit lens is quite noisy in early-middle layers. This corroborates evidence from prior work showing that image representations are aligned with language in early VLM layers \cite{featurevizaligned, latentlens}. 

\subsection{Object Confidence}\label{sec:confidence}
To quantify how well this approach recovers relevant semantic concepts from image tokens, we set up an object detection task using the COCO validation split \cite{COCO}, similar to the setup from \cite{jiang}. For each COCO category (e.g., ``backpack''), we sample $n$ images with that object and $n$ random images that do not contain that object. Let $o$ be the token for a particular category. 
For each image, we take the maximum $P(o)$ under verbalization lens across all image tokens as the model's confidence that an image contains $o$, focusing on one layer at a time.
We filter to include only single-token categories (56/80 COCO classes for Qwen3 models). As found in prior work \cite{jiang}, raw logit lens can distinguish COCO objects. 
But verbalization lens leads to much higher probabilities for images that contain $o$ (Figure~\ref{fig:detection}a). Verbalization lens also achieves consistent separation across layers, whereas raw logit lens only begins to work in late layers (Figure~\ref{fig:detection}b). Figure~\ref{fig:roc} shows similar results with ROC-AUC: raw logit lens offers some discrimination, but verbalization lens realizes near perfect AUC across layers.

\subsection{Object Localization}\label{sec:localization}

\begin{figure*}
    \centering
    \includegraphics[width=\linewidth]{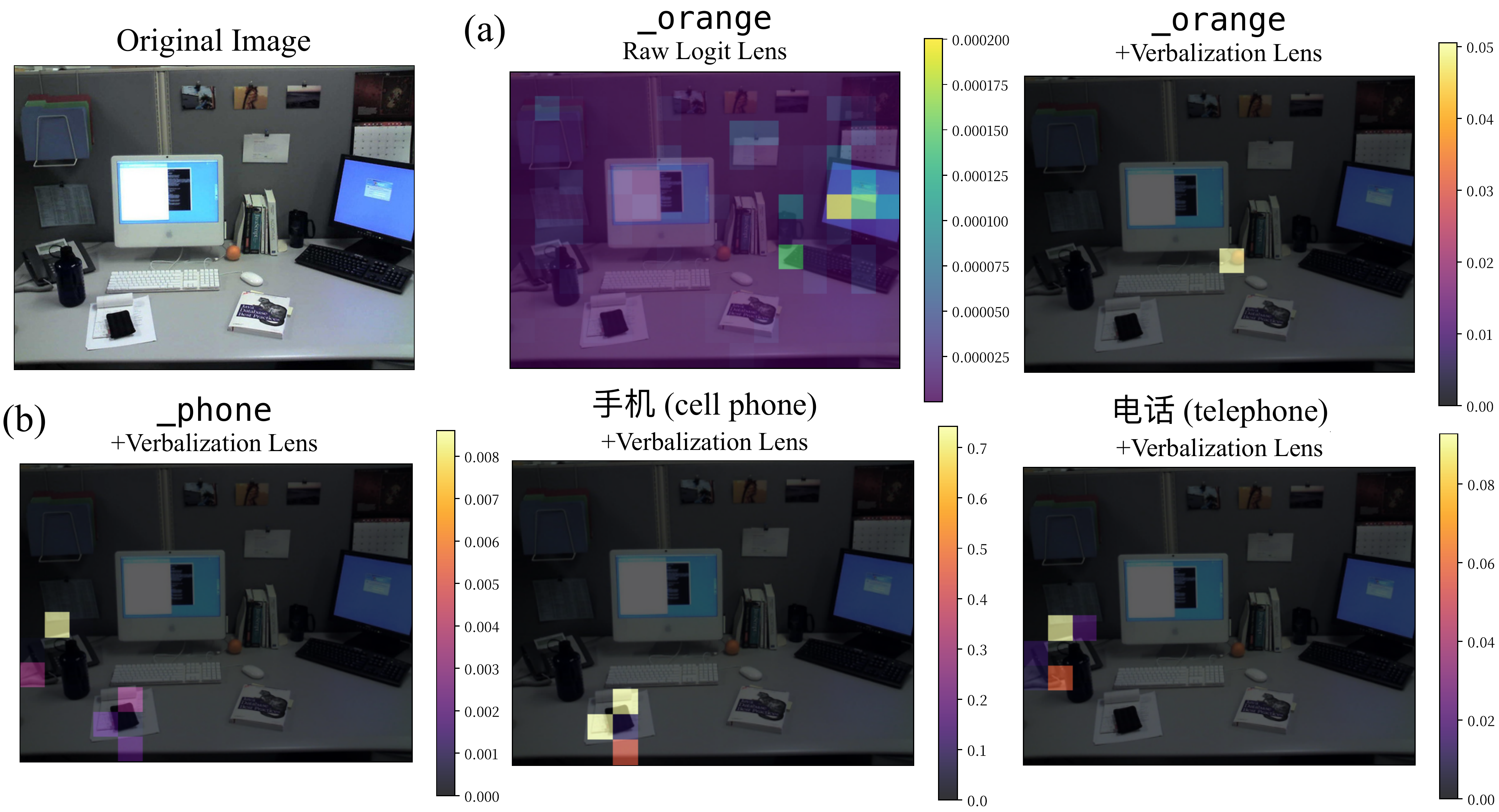}
    \caption{We use verbalization lens to measure $P(o)$ across image tokens for an object $o$, and find that high probabilities are concentrated in image tokens containing $o$. (a) $P(\text{orange})$=0.05 for a token containing an orange, whereas raw logit lens probabilities are lower and scattered across the image. (b) Localization is sensitive to the specific word used: $P(\text{phone})$ is high for both phones in the image, but ``\chinese{手机}'' (a Chinese word specifically meaning cell phone) localizes only the cell phone, and ``\chinese{电话}'' (a word mostly used for landlines) localizes only the landline. Results are for Qwen3-VL-2B (layer 15), with the top 10\% of verbalization heads.}
    \label{fig:localize_qual}
\end{figure*}

We find that verbalization lens generally reveals information localized to relevant regions of an image. Similar to Section~\ref{sec:confidence}, we calculate $P(o)$ across all image tokens at a given layer, which yields a heatmap of confidence scores across image tokens. We can use this approach to obtain probabilities at a particular location for any token in the model's vocabulary. For example, in Figure~\ref{fig:localize_qual}a, the token with the maximum $P(\text{orange})$ is the token containing a small orange. Localization is also sensitive to nuances in word meaning. While the token \texttt{\_phone} localizes image tokens that appear to contain a mobile or landline phone, ``\chinese{手机}'' (a Chinese word specifically meaning cell phone) localizes only the cell phone and ``\chinese{电话}'' (a word mostly used for landlines) localizes only the landline.

\begin{table}[t]
    \centering
    \caption{
        Token-level localization results for Qwen3-VL models. Here, we follow \cite{jiang} by taking the maximum probability for each token across layers. We use the top 10\% of verbalization heads based on results from Figure~\ref{fig:head-ablations}, and thus also sample 10\% of heads for the random baseline. For mIoU, we choose the best threshold based on 128 random images and report results for a different sample of 512 images, all from the COCO 2014 validation set. See Appendix~\ref{app:localization} for details. 
    }
    \label{tab:localization-allmodels}

    \begin{tabular}{lcccc}
        \toprule
        \multicolumn{1}{c}{} & \multicolumn{2}{c}{\textbf{Qwen3-VL-8B}} & \multicolumn{2}{c}{\textbf{Qwen3-VL-2B}} \\
        \midrule
        Method & mIoU & mAP & mIoU & mAP \\
        \midrule
        Raw Logit Lens  & 0.161 & 0.374 & 0.223 & 0.462  \\
        Random Heads    & 0.167 & 0.302 & 0.157 & 0.224  \\
        Verbalization Lens & 0.190 & 0.407 & 0.249 & 0.463  \\
        \bottomrule
    \end{tabular}
    
\end{table}

Following \cite{jiang}, we calculate segmentation metrics for our approach using the COCO 2014 validation set \cite{COCO}. 
For a given image, we retrieve all of the single-token categories\footnote{This number is roughly similar across models; see Table~\ref{tab:models}.} $o$ in that image and use our lens to generate segmentation maps for each category in the image. To generate a segmentation map, we obtain a score $P(o)$ for each token by taking the maximum probability of $o$ across layers. 
We then binarize these probabilities to score lens predictions against ground truth masks by choosing a threshold $\tau$ on a disjoint set of images (or sweep across $\tau$ for AP), and average resulting scores across all objects/images. 
Unlike prior object localization work \cite{jiang, gandelsman2024clip, chefer2021}, we evaluate at the token-scale instead of upsampling our predictions to the pixel scale.\footnote{We define a patch as containing an object if more than 10\% of its pixels correspond to the ground-truth COCO mask.}

We compare to logit lens and an equivalent number of random heads.
Table~\ref{tab:localization-allmodels} shows that verbalization lens is slightly more spatially localized than baselines. However, numbers are overall quite low---this may in part be due to register token behavior \cite{darcet2024register, jiangdravidregister2025}, where general information about an image is stored in arbitrary tokens (e.g., Figure~\ref{fig:lens_qual}f).

\subsection{Comparison to LatentLens}\label{sec:comparing-latentlens}

LatentLens \citep{latentlens} is an alternative approach to decoding image activations: instead of using unembeddings, it measures cosine similarity of image tokens against a large pool of contextual text embeddings from intermediate LLM layers.
However, \citet{latentlens} report failure modes in the early and mid-layers of some larger models such as Llava-Next-34B.
Adopting the same LLM judge interpretability metric as \citet{latentlens}, we find that verbalization lens produces interpretable tokens even in these difficult settings, with a 40.5 point increase in judge scores for Llava-Next-34B on average across layers. For all other models, our approach is on par with LatentLens; see Appendix~\ref{app:latentlens-details} for details. 

\section{Editing Conceptual Information}\label{sec:editing}

Are the representations read out by these verbalization heads a mirage, or are they causally relevant for the model's ``understanding'' of an image? Instead of merely verbalizing hidden state representations, we use the subspace read by $\mathbf{L}_p$ to edit Qwen3-VL models' \emph{perception} of an image, finding that we can, e.g., cause the model to describe an ant in an image as an iPod. 

\begin{figure*}
    \centering
    \includegraphics[width=\linewidth]{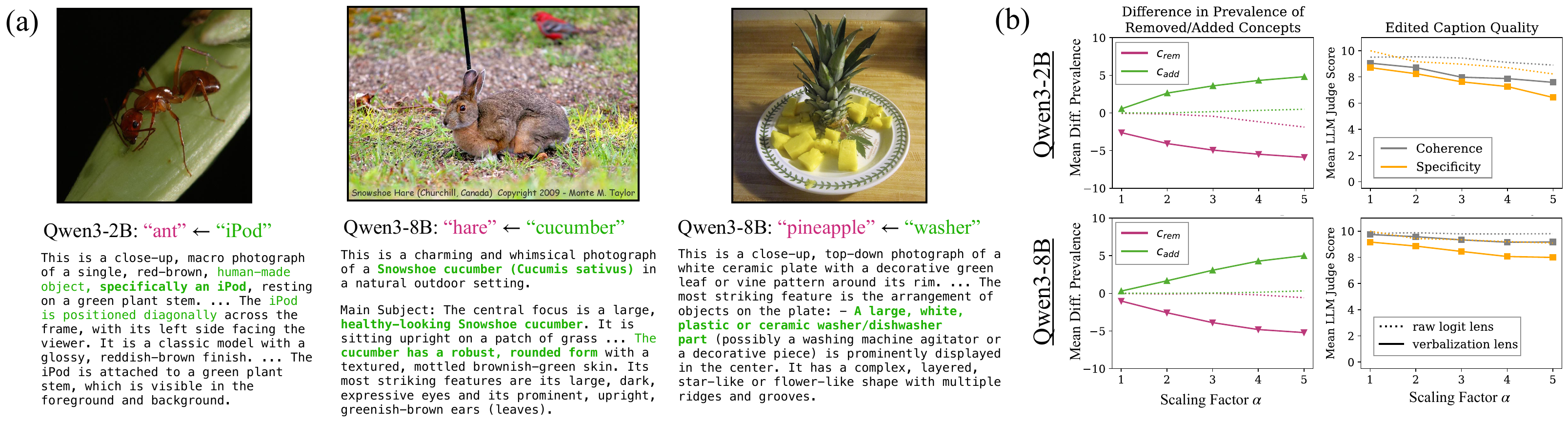}
    \caption{We can use our verbalization transformation to edit image activations, e.g., replacing the concept of ``ant'' with ``iPod.'' (a) Qualitative examples for editing objects within an image. Rank is constant based on our initial sweep; we show $\alpha$=4 for the first two images, and $\alpha$=5 for the third.
    (b) Quantitative LLM judge scores across random ImageNet \cite{imagenet} images ($n$=256). Verbalization lens yields latent vectors that can effectively edit image representations without heavily damaging irrelevant concepts, or degrading caption coherence. Edits are most effective with a larger scaling factor $\alpha$. We use a rank for each model that explains the top $\rho$=60\% of verbalization energy, based on an initial sweep over 100 images (Appendix~\ref{app:edit_sweep}).}
    \label{fig:edit_main}
\end{figure*}

\subsection{Approach}

Let $c_{\text{rem}}$ correspond to a concept we want to remove, and $c_{\text{add}}$ be a concept we want to add in its place. Here, we focus on single-token concepts. If we take the unembedding vector $\mathbf{u}_c\in\mathbb{R}^{d_{\text{model}}}$ corresponding to the token for the concept $c$, we can obtain a latent vector $\mathbf{v}_c$ for that concept using the inverse of the transformation from Equation~\ref{eq:build_ocr}: 

\begin{equation}
    \mathbf{v}_c = \mathbf{L}_p^{-1}\mathbf{u}_c.
\end{equation}

\noindent where $\mathbf{v}_c\in\mathbb{R}^{d_{\text{model}}}$ is a direction in model activation space that encodes the concept $c$. Although $\mathbf{L}_p$ is typically full-rank (and thus invertible), we find that in practice it has a long tail of small singular values (e.g., Figure~\ref{fig:svs_qwen2b}). This means that directly taking the inverse of $\mathbf{L}_p$ would cause an explosion of singular values in $\mathbf{L}_p^{-1}$. Instead of taking the full inverse, we can take a pseudo-inverse:

\begin{align}
    \mathbf{L}_p &= \mathbf{U}\mathbf{\Sigma}\mathbf{V}^*\\
    \mathbf{L}_{p,k}^{+} &= \mathbf{V}[:,:k]\mathbf{\Sigma}^{-1}\mathbf{U}^*[:k]
\end{align}

\noindent where $1\leq k\leq d_{\text{model}}$ is an integer hyperparameter, chosen via sweep on a held-out set. 
Here, we interpret bottom singular values as noise, manually setting them to zero rather than letting them affect our transformation. This approach implicitly views $\mathbf{L}_p$ as reading from a low-rank subspace of the residual stream, as described in previous work \cite{lre}. It also means that our edit is more surgical, as we are only affecting activations within a low-rank subspace.

Once we have obtained our latent vectors $\mathbf{v}_{c_{\text{rem}}},\mathbf{v}_{c_{\text{add}}}$ for the concepts we would like to add and remove, we edit activations $\mathbf{a}^{(l)}_{t}\in\mathbb{R}^{d_{\text{model}}}$ across all layers $l$ and image token positions $t$. For a given activation vector, we remove the component of that activation in the direction of $\mathbf{v}_{c_{\text{rem}}}$, and add in $\mathbf{v}_{c_{\text{add}}}$ at the same relative magnitude:

\begin{align}
    \lambda &= (\mathbf{a}^{(l)}_t\cdot \mathbf{v}_{c_{\text{rem}}})  / ||\mathbf{v}_{c_{\text{rem}}}||^2 \\
    \mathbf{a}^{(l)*}_t &= \mathbf{a}^{(l)}_t - \lambda\mathbf{v}_{c_{\text{rem}}} + \alpha\lambda\mathbf{v}_{c_{\text{add}}}\label{eq:editing}
\end{align}

\noindent where we allow for a scaling factor $\alpha\in\mathbb{R}$ on the added concept $c_{\text{add}}$ as an additional hyperparameter.
Note that if $c_{\text{rem}}$ is not present in this activation vector, $\lambda$ will be close to zero, and Equation~\ref{eq:editing} becomes a no-op. This is by design: we do not want to edit token positions where there is nothing to replace. Thus, while we apply this edit to all image tokens, it only has an effect at relevant token positions.

\subsection{Evaluation}\label{sec:edit_eval}

To evaluate the efficacy of our edits, we prompt models to free generate descriptions with intervened image representations. We prompt o4-mini to score the generated captions according to four metrics on a scale from 0 to 10:
\begin{itemize}
    \item \textbf{Prevalence of $c_{\text{add}}$.} e.g., how prominently does the concept of `dog' feature in this caption?
    \item \textbf{Prevalence of $c_{\text{rem}}$.} e.g., how prominently does the concept of `motorcycle' feature in this caption?
    \item \textbf{Specificity.} Apart from any differences regarding the concepts $c_{\text{rem}}$ and $c_{\text{add}}$, how similar is the counterfactual caption to the ground truth vanilla caption?
    \item \textbf{Coherence.} In terms of writing quality, how coherent is this caption? 
\end{itemize}

See Appendix~\ref{app:appendix-llm-judge} for full prompts. We obtain scores for all generated and vanilla captions. To measure edit success, we measure the $\Delta$ in prevalence scores for a counterfactual caption compared to scores for the vanilla generation. 
We evaluate on ImageNet \cite{imagenet} images as they are generally focused on one subject at a time: images are assigned $c_{\text{rem}}$ based on their class label, sampling only from images with single-token class labels. Then, $c_{\text{add}}$ is randomly sampled from all other single-token ImageNet class names.

\subsection{Results}\label{sec:edit_results}

\begin{figure*}[!t]
    \centering
    \includegraphics[width=\linewidth]{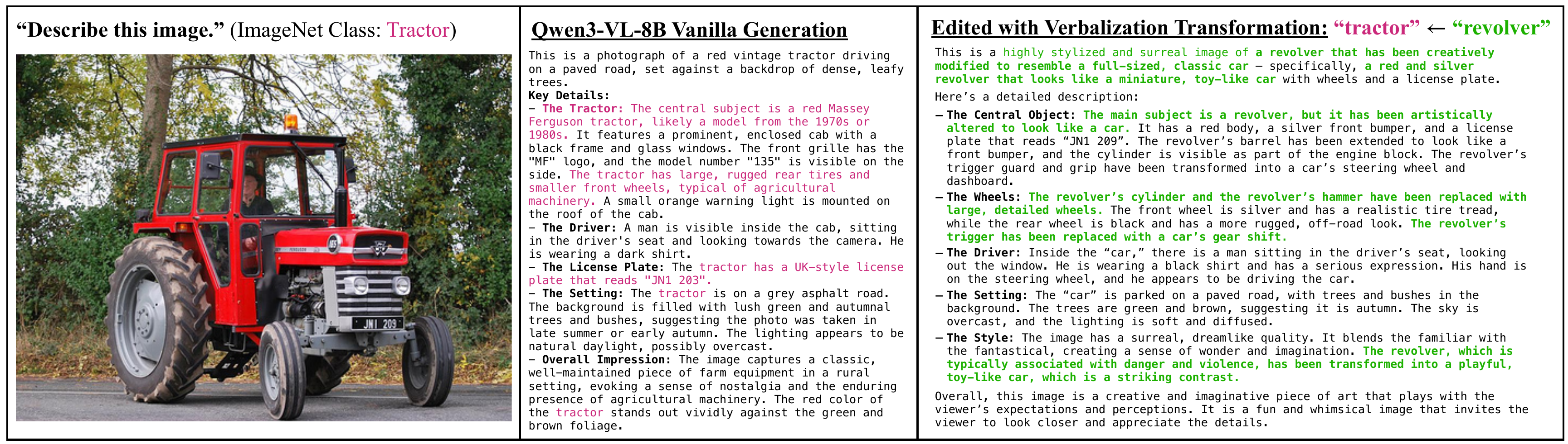}
    \caption{We can use our verbalization transformation to edit Qwen3-VL-8B image activations, replacing the concept of ``tractor'' with ``revolver.'' We obtain latent vectors for these concepts using the inverse of our verbalization matrix. After subtracting the latent vector for ``tractor'' and adding the vector for ``revolver,'' Qwen3-VL-8B describes the main subject of the image as a revolver. We leave all other image features untouched. The model reconciles remaining visual features (like wheels, a license plate, and a driver) by describing the image as a ``surreal image of a revolver that has been creatively modified to resemble a full-sized, classic car.'' Both captions are generated with greedy decoding. This generation uses $\alpha$=4; across all examples, we use the top $\rho$=60\% of $\mathbf{L}_p$ energy, where $p$=10\%. }
    \label{fig:edit_qual}
\end{figure*}

Editing with latent vectors obtained via the verbalization head subspace is more effective than directly editing with unembeddings. After performing an initial sweep over rank and $\alpha$ (see Appendix~\ref{app:edit_sweep} for details), we select the rank that explains 60\% of the energy of $\mathbf{L}_p$ and run our edit with this setting for $n$=256 random images, with results shown in Figure~\ref{fig:edit_main}b. We find using $\alpha>1$ is most effective. Intuitively, this means that $c_{\text{add}}$ must be added more strongly than $c_{\text{rem}}$ was present in order to have a causal effect; perhaps this is to override remaining information in the residual stream related to $c_{\text{rem}}$ but not captured in $\mathbf{v}_{c_\text{rem}}$. 

We show three abridged qualitative examples in Figure~\ref{fig:edit_main}a. 
We can see that, e.g., the concept of ``ant'' has been cleanly replaced by the concept of ``iPod,'' while other visual features (like background descriptions) remain intact. These edits preserve low-level visual features of the original object: i.e., the color and texture of the ant is now applied to the iPod, with the model describing the iPod as being ``a classic model with a glossy, reddish-brown finish.'' 

Figure~\ref{fig:edit_qual} shows a full example edit for Qwen3-VL-8B alongside the model's vanilla generation. Here, $c_{\text{rem}}$=``tractor,'' and we randomly sample $c_{\text{add}}$=``revolver''. Our edit is successful: the model does not describe the image as containing a tractor, but rather as an image of a revolver. Interestingly, because other aspects of the image remain untouched (wheels, license plate, bumper, and man inside the tractor), Qwen synthesizes these attributes into a ``surreal'' description of a ``revolver that has been creatively modified to resemble a full-sized, classic car.''

\section{Related Work}

\paragraph{Lenses.} Logit lens \cite{nostalgebraist} was the first to find that LLM internal hidden states can be projected to vocabulary directions with interpretable results, and was followed by a large body of work on other ``lenses'' to decode internal states \cite{nostalgebraist, tunedlens, lre, pal2023future, jumptoconclusions, geva2022transformerfeedforwardlayersbuild, katz2024backwardlensprojectinglanguage, feucht2025dualroute, feucht2025arithmetic, gurnee2026verbalizable}. Prior work has applied logit lens approaches to VLMs \cite{neo, jiang, li2025reading}; one approach \cite{latentlens} provides an alternative to vocabulary decoding using cached latent representations of text. Earlier, \cite{joseph2024laying} also showed that CLIP's classifier head can act as a ``logit lens'' for its intermediate states. Our work augments the interpretability of these vanilla logit lens approaches. 

\paragraph{Language and Vision.} Previous work studies ``multimodal'' neurons in CLIP that respond to both text and images referring to one concept \cite{goh2021multimodal, joanna}. With just a small adapter module, one can map from image to text \cite{clipcap, tsimpoukelli2021, manas-etal-2023-mapl}, even with both models frozen \cite{merullo2023, schwettmann2023}. Other works study vision representations through the lens of language \cite{gandelsman2024clip, chen2023interpretingcontrollingvisionfoundation}, or differentiation between vision and language representations \cite{papadimitriou2025, sourcemodality, platonic}. Language priors can conflict with visual features \cite{ethaconflicting2025, shu2025semanticsmisleadvisionmitigating}, particularly for OCR \cite{karamolegkou2026readingguessingvisualgrounding,
liang2026visualmeritlinguisticcrutch,
he2025seeingbelievingmitigatingocr}.
Our results are evocative of recent work on how VLMs use language to mediate visual understanding \cite{vlmsneedwords,tartaglini2025bottlenecks,paraselli2026slowseeslowsuppress}. While earlier work claims that alignment arises only in middle layers \cite{jiang2024hallucinationaugmentedcontrastivelearning, schwettmann2023, venhoff2025toolatetorecall, neo}, others assert early-layer alignment \cite{featurevizaligned, latentlens}, which our method also shows. 

\paragraph{Other heads.} We note conceptual overlap between verbalization heads and two types of attention heads studied in prior work: \emph{filter heads} \cite{sensharma2023filter} and \emph{gaze heads} \cite{gandikota2026gazeheads}. We compare against these and find modest overlap in Appendix~\ref{app:overlap}, suggesting that there may be a relationship between these components. This further supports our claim that the narrow setting of OCR provides a straightforward way of identifying heads responsible for more general VLM mechanisms.

\section{Conclusion}

We find that attention heads responsible for OCR can be used to verbalize general semantics of non-text image tokens across four VLMs. We collapse the weights of these attention heads into a single matrix that improves upon logit lens, and show that the subspace this matrix reads from is causally relevant for model descriptions of images. Our results show how study of narrow mechanisms can shed light on broader interpretability problems.

\section*{Acknowledgements}

SF thanks Si Wu for discussions on language and letters throughout this project, as well as Rohit Gandikota and Arnab Sen Sharma for feedback on early experiments. We thank Andy Arditi, Eric Todd, and Grace Proebsting for feedback on initial drafts. SF is funded by a grant from Coefficient Giving. BK and DB are funded by NSF \#2403304.


{
    \small
    \bibliographystyle{ieeenat_fullname}
    \bibliography{main}

@String(CVPR= {IEEE Conf. Comput. Vis. Pattern Recog.})

@String(ICCV= {Int. Conf. Comput. Vis.})

@String(ECCV= {Eur. Conf. Comput. Vis.})

@String(ICLR = {Int. Conf. Learn. Represent.})

@String(CVPR  = {CVPR})

@String(ICCV  = {ICCV})

@String(ECCV  = {ECCV})

@String(ICLR  = {ICLR})

@inproceedings{
    feucht2025dualroute,
    title={The Dual-Route Model of Induction},
    author={Sheridan Feucht and Eric Todd and Byron Wallace and David Bau},
    booktitle={Second Conference on Language Modeling},
    year={2025},
    url={https://arxiv.org/abs/2504.03022}
}

@inproceedings{feucht2025arithmetic,
  title={Vector Arithmetic in Concept and Token Subspaces},
  author={Sheridan Feucht and Byron Wallace and David Bau},
  booktitle={Second Mechanistic Interpretability Workshop at NeurIPS},
  year={2025},
  url={https://arithmetic.baulab.info}
}

@inproceedings{COCO,
  author       = {Tsung{-}Yi Lin and
                  Michael Maire and
                  Serge J. Belongie and
                  James Hays and
                  Pietro Perona and
                  Deva Ramanan and
                  Piotr Doll{\'{a}}r and
                  C. Lawrence Zitnick},
  editor       = {David J. Fleet and
                  Tom{\'{a}}s Pajdla and
                  Bernt Schiele and
                  Tinne Tuytelaars},
  title        = {Microsoft {COCO:} Common Objects in Context},
  booktitle    = {Computer Vision - {ECCV} 2014 - 13th European Conference, Zurich,
                  Switzerland, September 6-12, 2014, Proceedings, Part {V}},
  series       = {Lecture Notes in Computer Science},
  volume       = {8693},
  pages        = {740--755},
  publisher    = {Springer},
  year         = {2014},
  url          = {https://doi.org/10.1007/978-3-319-10602-1\_48},
  doi          = {10.1007/978-3-319-10602-1\_48},
  bibsource    = {dblp computer science bibliography, https://dblp.org}
}

@misc{jiang,
      title={Interpreting and Editing Vision-Language Representations to Mitigate Hallucinations}, 
      author={Nick Jiang and Anish Kachinthaya and Suzie Petryk and Yossi Gandelsman},
      year={2025},
      eprint={2410.02762},
      archivePrefix={arXiv},
      primaryClass={cs.CV},
      url={https://arxiv.org/abs/2410.02762}, 
}

@inproceedings{chefer2021,
  title     = {Transformer Interpretability Beyond Attention Visualization},
  author    = {Chefer, Hila and Gur, Shir and Wolf, Lior},
  booktitle = {Proceedings of the IEEE/CVF Conference on Computer Vision and Pattern Recognition},
  pages     = {782--791},
  year      = {2021}
}

@article{goh2021multimodal,
  author = {Goh, Gabriel and †, Nick Cammarata and †, Chelsea Voss and Carter, Shan and Petrov, Michael and Schubert, Ludwig and Radford, Alec and Olah, Chris},
  title = {Multimodal Neurons in Artificial Neural Networks},
  journal = {Distill},
  year = {2021},
  note = {https://distill.pub/2021/multimodal-neurons},
  doi = {10.23915/distill.00030}
}

@inproceedings{joanna,
  author       = {Joanna Materzynska and
                  Antonio Torralba and
                  David Bau},
  title        = {Disentangling visual and written concepts in {CLIP}},
  booktitle    = {{IEEE/CVF} Conference on Computer Vision and Pattern Recognition,
                  {CVPR} 2022, New Orleans, LA, USA, June 18-24, 2022},
  pages        = {16389--16398},
  publisher    = {{IEEE}},
  year         = {2022},
  url          = {https://doi.org/10.1109/CVPR52688.2022.01592},
  doi          = {10.1109/CVPR52688.2022.01592},
  bibsource    = {dblp computer science bibliography, https://dblp.org}
}

@inproceedings{darcet2024register,
  author       = {Timoth{\'{e}}e Darcet and
                  Maxime Oquab and
                  Julien Mairal and
                  Piotr Bojanowski},
  title        = {Vision Transformers Need Registers},
  booktitle    = {The Twelfth International Conference on Learning Representations,
                  {ICLR} 2024, Vienna, Austria, May 7-11, 2024},
  publisher    = {OpenReview.net},
  year         = {2024},
  url          = {https://openreview.net/forum?id=2dnO3LLiJ1},
  bibsource    = {dblp computer science bibliography, https://dblp.org}
}

@inproceedings{jiangdravidregister2025,
  author       = {Nick Jiang and
                  Amil Dravid and
                  Alexei A. Efros and
                  Yossi Gandelsman},
  editor       = {Danielle Belgrave and
                  Cheng Zhang and
                  Laura N. Montoya and
                  Hsuan{-}Tien Lin and
                  Razvan Pascanu and
                  Piotr Koniusz and
                  Marzyeh Ghassemi and
                  Nancy Chen and
                  Iv{\'{a}}n Vladimir Meza Ru{\'{\i}}z and
                  Arturo Loaiza{-}Bonilla},
  title        = {Vision Transformers Don't Need Trained Registers},
  booktitle    = {Advances in Neural Information Processing Systems 38: Annual Conference
                  on Neural Information Processing Systems 2025, NeurIPS 2025, San Diego,
                  CA, USA, December 2-7, 2025 / Mexico City, Mexico, November 30 - December
                  5, 2025},
  year         = {2025},
  url          = {http://papers.nips.cc/paper\_files/paper/2025/hash/51bae6441380c4c4cab87f430be89ac2-Abstract-Conference.html},
  bibsource    = {dblp computer science bibliography, https://dblp.org}
}

@inproceedings{
latentlens,
title={LatentLens: Revealing Highly Interpretable Visual Tokens in {LLM}s},
author={Benno Krojer and Shravan Nayak and Oscar Ma{\~n}as and Vaibhav Adlakha and Desmond Elliott and Siva Reddy and Marius Mosbach},
booktitle={Forty-third International Conference on Machine Learning},
year={2026},
url={https://openreview.net/forum?id=qJFatzEGQc}
}

@misc{featurevizaligned,
      title={Representations of Text and Images Align From Layer One}, 
      author={Evžen Wybitul and Javier Rando and Florian Tramèr and Stanislav Fort},
      year={2026},
      eprint={2601.08017},
      archivePrefix={arXiv},
      primaryClass={cs.CV},
      url={https://arxiv.org/abs/2601.08017}, 
}

@misc{jiang2024hallucinationaugmentedcontrastivelearning,
      title={Hallucination Augmented Contrastive Learning for Multimodal Large Language Model}, 
      author={Chaoya Jiang and Haiyang Xu and Mengfan Dong and Jiaxing Chen and Wei Ye and Ming Yan and Qinghao Ye and Ji Zhang and Fei Huang and Shikun Zhang},
      year={2024},
      eprint={2312.06968},
      archivePrefix={arXiv},
      primaryClass={cs.CV},
      url={https://arxiv.org/abs/2312.06968}, 
}

@article{imagenet,
  author       = {Olga Russakovsky and
                  Jia Deng and
                  Hao Su and
                  Jonathan Krause and
                  Sanjeev Satheesh and
                  Sean Ma and
                  Zhiheng Huang and
                  Andrej Karpathy and
                  Aditya Khosla and
                  Michael S. Bernstein and
                  Alexander C. Berg and
                  Li Fei{-}Fei},
  title        = {ImageNet Large Scale Visual Recognition Challenge},
  journal      = {CoRR},
  volume       = {abs/1409.0575},
  year         = {2014},
  url          = {http://arxiv.org/abs/1409.0575},
  eprinttype   = {arXiv},
  eprint       = {1409.0575},
  bibsource    = {dblp computer science bibliography, https://dblp.org}
}

@article{sensharma2023filter,
    title={LLMs Process Lists With General Filter Heads}, 
    author={Arnab Sen Sharma and Giordano Rogers and Natalie Shapira and David Bau},
    journal={The Fourteenth International Conference on Learning Representations (ICLR 2026)},
    year={2025},
    eprint={2510.26784},
    archivePrefix={arXiv},
    primaryClass={cs.CL}
}

@article{gandikota2026gazeheads,
  title={Gaze Heads: How VLMs Look at What They Describe},
  author={Rohit Gandikota and David Bau},
  journal={arXiv preprint arXiv:2606.14703},
  year={2026}
}

@InProceedings{balanced_vqa_v2,
author = {Yash Goyal and Tejas Khot and Douglas Summers{-}Stay and Dhruv Batra and Devi Parikh},
title = {Making the {V} in {VQA} Matter: Elevating the Role of Image Understanding in {V}isual {Q}uestion {A}nswering},
booktitle = {Conference on Computer Vision and Pattern Recognition (CVPR)},
year = {2017},
}

@article{dehaenereview,
	author = {Dehaene, Stanislas and Cohen, Laurent and Morais, Jos{\'e} and Kolinsky, R{\'e}gine},
	date = {2015/04/01},
	doi = {10.1038/nrn3924},
	id = {Dehaene2015},
	isbn = {1471-0048},
	journal = {Nature Reviews Neuroscience},
	number = {4},
	pages = {234--244},
	title = {Illiterate to literate: behavioural and cerebral changes induced by reading acquisition},
	url = {https://doi.org/10.1038/nrn3924},
	volume = {16},
	year = {2015}}

@misc{liu2024llavanext,
    title={LLaVA-NeXT: Improved reasoning, OCR, and world knowledge},
    url={https://llava-vl.github.io/blog/2024-01-30-llava-next/},
    author={Liu, Haotian and Li, Chunyuan and Li, Yuheng and Li, Bo and Zhang, Yuanhan and Shen, Sheng and Lee, Yong Jae},
    month={January},
    year={2024}
}

@article{molmo2,
  author       = {Christopher Clark and
                  Jieyu Zhang and
                  Zixian Ma and
                  Jae Sung Park and
                  Mohammadreza Salehi and
                  Rohun Tripathi and
                  Sangho Lee and
                  Zhongzheng Ren and
                  Chris Dongjoo Kim and
                  Yinuo Yang and
                  Vincent Shao and
                  Yue Yang and
                  Weikai Huang and
                  Ziqi Gao and
                  Taira Anderson and
                  Jianrui Zhang and
                  Jitesh Jain and
                  George Stoica and
                  Winson Han and
                  Ali Farhadi and
                  Ranjay Krishna},
  title        = {Molmo2: Open Weights and Data for Vision-Language Models with Video
                  Understanding and Grounding},
  journal      = {CoRR},
  volume       = {abs/2601.10611},
  year         = {2026},
  url          = {https://doi.org/10.48550/arXiv.2601.10611},
  doi          = {10.48550/ARXIV.2601.10611},
  eprinttype   = {arXiv},
  eprint       = {2601.10611},
  bibsource    = {dblp computer science bibliography, https://dblp.org}
}

@misc{bai2025qwen3vltechnicalreport,
      title={Qwen3-VL Technical Report}, 
      author={Shuai Bai and Yuxuan Cai and Ruizhe Chen and Keqin Chen and Xionghui Chen and Zesen Cheng and Lianghao Deng and Wei Ding and Chang Gao and Chunjiang Ge and Wenbin Ge and Zhifang Guo and Qidong Huang and Jie Huang and Fei Huang and Binyuan Hui and Shutong Jiang and Zhaohai Li and Mingsheng Li and Mei Li and Kaixin Li and Zicheng Lin and Junyang Lin and Xuejing Liu and Jiawei Liu and Chenglong Liu and Yang Liu and Dayiheng Liu and Shixuan Liu and Dunjie Lu and Ruilin Luo and Chenxu Lv and Rui Men and Lingchen Meng and Xuancheng Ren and Xingzhang Ren and Sibo Song and Yuchong Sun and Jun Tang and Jianhong Tu and Jianqiang Wan and Peng Wang and Pengfei Wang and Qiuyue Wang and Yuxuan Wang and Tianbao Xie and Yiheng Xu and Haiyang Xu and Jin Xu and Zhibo Yang and Mingkun Yang and Jianxin Yang and An Yang and Bowen Yu and Fei Zhang and Hang Zhang and Xi Zhang and Bo Zheng and Humen Zhong and Jingren Zhou and Fan Zhou and Jing Zhou and Yuanzhi Zhu and Ke Zhu},
      year={2025},
      eprint={2511.21631},
      archivePrefix={arXiv},
      primaryClass={cs.CV},
      url={https://arxiv.org/abs/2511.21631}, 
}

@misc{yang2025qwen3technicalreport,
      title={Qwen3 Technical Report}, 
      author={An Yang and Anfeng Li and Baosong Yang and Beichen Zhang and Binyuan Hui and Bo Zheng and Bowen Yu and Chang Gao and Chengen Huang and Chenxu Lv and Chujie Zheng and Dayiheng Liu and Fan Zhou and Fei Huang and Feng Hu and Hao Ge and Haoran Wei and Huan Lin and Jialong Tang and Jian Yang and Jianhong Tu and Jianwei Zhang and Jianxin Yang and Jiaxi Yang and Jing Zhou and Jingren Zhou and Junyang Lin and Kai Dang and Keqin Bao and Kexin Yang and Le Yu and Lianghao Deng and Mei Li and Mingfeng Xue and Mingze Li and Pei Zhang and Peng Wang and Qin Zhu and Rui Men and Ruize Gao and Shixuan Liu and Shuang Luo and Tianhao Li and Tianyi Tang and Wenbiao Yin and Xingzhang Ren and Xinyu Wang and Xinyu Zhang and Xuancheng Ren and Yang Fan and Yang Su and Yichang Zhang and Yinger Zhang and Yu Wan and Yuqiong Liu and Zekun Wang and Zeyu Cui and Zhenru Zhang and Zhipeng Zhou and Zihan Qiu},
      year={2025},
      eprint={2505.09388},
      archivePrefix={arXiv},
      primaryClass={cs.CL},
      url={https://arxiv.org/abs/2505.09388}, 
}

@article{elhage2021mathematical,
   title={A Mathematical Framework for Transformer Circuits},
   author={Elhage, Nelson and Nanda, Neel and Olsson, Catherine and Henighan, Tom and Joseph, Nicholas and Mann, Ben and Askell, Amanda and Bai, Yuntao and Chen, Anna and Conerly, Tom and DasSarma, Nova and Drain, Dawn and Ganguli, Deep and Hatfield-Dodds, Zac and Hernandez, Danny and Jones, Andy and Kernion, Jackson and Lovitt, Liane and Ndousse, Kamal and Amodei, Dario and Brown, Tom and Clark, Jack and Kaplan, Jared and McCandlish, Sam and Olah, Chris},
   year={2021},
   journal={Transformer Circuits Thread},
   note={https://transformer-circuits.pub/2021/framework/index.html}
}

@inproceedings{ladstages2025,
  author       = {Vedang Lad and
                  Jin Hwa Lee and
                  Wes Gurnee and
                  Max Tegmark},
  editor       = {Danielle Belgrave and
                  Cheng Zhang and
                  Laura N. Montoya and
                  Hsuan{-}Tien Lin and
                  Razvan Pascanu and
                  Piotr Koniusz and
                  Marzyeh Ghassemi and
                  Nancy Chen and
                  Iv{\'{a}}n Vladimir Meza Ru{\'{\i}}z and
                  Arturo Loaiza{-}Bonilla},
  title        = {Remarkable Robustness of LLMs: Stages of Inference?},
  booktitle    = {Advances in Neural Information Processing Systems 38: Annual Conference
                  on Neural Information Processing Systems 2025, NeurIPS 2025, San Diego,
                  CA, USA, December 2-7, 2025 / Mexico City, Mexico, November 30 - December
                  5, 2025},
  year         = {2025},
  url          = {http://papers.nips.cc/paper\_files/paper/2025/hash/bcad07d4bfab51243efaa08b8ed475b3-Abstract-Conference.html},
  bibsource    = {dblp computer science bibliography, https://dblp.org}
}

@inproceedings{todd2024function,
    title={Function Vectors in Large Language Models}, 
    author={Eric Todd and Millicent L. Li and Arnab Sen Sharma and Aaron Mueller and Byron C. Wallace and David Bau},
    booktitle={The Twelfth International Conference on Learning Representations},
    url={https://openreview.net/forum?id=AwyxtyMwaG},
    note={arXiv:2310.15213},
    year={2024},
}

@misc{joseph2024laying,
  title={Laying the Foundations for Vision and Multimodal Mechanistic Interpretability \& Open Problems},
  author={Sonia Joseph and Neel Nanda},
  year={2024},
  url={https://www.alignmentforum.org/posts/kobJymvvcvhbjWFKe/laying-the-foundations-for-vision-and-multimodal-mechanistic},
  note={Accessed: 2024-06-28}
}

@inproceedings{gandelsman2024clip,
  author       = {Yossi Gandelsman and
                  Alexei A. Efros and
                  Jacob Steinhardt},
  title        = {Interpreting CLIP's Image Representation via Text-Based Decomposition},
  booktitle    = {The Twelfth International Conference on Learning Representations,
                  {ICLR} 2024, Vienna, Austria, May 7-11, 2024},
  publisher    = {OpenReview.net},
  year         = {2024},
  url          = {https://openreview.net/forum?id=5Ca9sSzuDp},
  bibsource    = {dblp computer science bibliography, https://dblp.org}
}

@misc{nostalgebraist,
  title={Interpreting GPT: The Logit Lens},
  author={Nostalgebraist},
  year={2020},
  month={Aug},
  howpublished={\url{https://www.alignmentforum.org/posts/AcKRB8wDpdaN6v6ru/interpreting-gpt-the-logit-lens}},
  note={Accessed: 23 Sep 2024}
}

@article{neo,
  title={Towards Interpreting Visual Information Processing in Vision-Language Models},
  author={Clement Neo and Luke Ong and Philip H. S. Torr and Mor Geva and David Krueger and Fazl Barez},
  journal={ArXiv},
  year={2024},
  volume={abs/2410.07149},
  url={https://api.semanticscholar.org/CorpusID:273233293}
}

@article{li2025reading,
  title={Reading Images Like Texts: Sequential Image Understanding in Vision-Language Models},
  author={Li, Yueyan and Zhao, Chenggong and Zang, Zeyuan and Yuan, Caixia and Wang, Xiaojie},
  journal={arXiv preprint arXiv:2509.19191},
  year={2025}
}

@article{andrewleeqk2026,
  author       = {Andrew Lee and
                  Yonatan Belinkov and
                  Fernanda B. Vi{\'{e}}gas and
                  Martin Wattenberg},
  title        = {Decomposing Query-Key Feature Interactions Using Contrastive Covariances},
  journal      = {CoRR},
  volume       = {abs/2602.04752},
  year         = {2026},
  url          = {https://doi.org/10.48550/arXiv.2602.04752},
  doi          = {10.48550/ARXIV.2602.04752},
  eprinttype   = {arXiv},
  eprint       = {2602.04752},
  bibsource    = {dblp computer science bibliography, https://dblp.org}
}

@inproceedings{lre,
    title={Linearity of Relation Decoding in Transformer Language Models}, 
    author={Evan Hernandez and Arnab Sen Sharma and Tal Haklay and Kevin Meng and Martin Wattenberg and Jacob Andreas and Yonatan Belinkov and David Bau},
    booktitle={Proceedings of the 2024 International Conference on Learning Representations},
    year={2024},
}

@inproceedings{pal2023future,
    title={Future Lens: Anticipating Subsequent Tokens from a Single Hidden State},
    author={Pal, Koyena and Sun, Jiuding and Yuan, Andrew and Wallace, Byron C and Bau, David},
    booktitle={Proceedings of the 27th Conference on Computational Natural Language Learning (CoNLL)},
    pages={548--560},
    year={2023}
}

@article{tunedlens,
  author       = {Nora Belrose and
                  Zach Furman and
                  Logan Smith and
                  Danny Halawi and
                  Igor Ostrovsky and
                  Lev McKinney and
                  Stella Biderman and
                  Jacob Steinhardt},
  title        = {Eliciting Latent Predictions from Transformers with the Tuned Lens},
  journal      = {CoRR},
  volume       = {abs/2303.08112},
  year         = {2023},
  url          = {https://doi.org/10.48550/arXiv.2303.08112},
  doi          = {10.48550/ARXIV.2303.08112},
  eprinttype   = {arXiv},
  eprint       = {2303.08112},
  bibsource    = {dblp computer science bibliography, https://dblp.org}
}

@inproceedings{jumptoconclusions,
  author       = {Alexander Yom Din and
                  Taelin Karidi and
                  Leshem Choshen and
                  Mor Geva},
  editor       = {Nicoletta Calzolari and
                  Min{-}Yen Kan and
                  V{\'{e}}ronique Hoste and
                  Alessandro Lenci and
                  Sakriani Sakti and
                  Nianwen Xue},
  title        = {Jump to Conclusions: Short-Cutting Transformers with Linear Transformations},
  booktitle    = {Proceedings of the 2024 Joint International Conference on Computational
                  Linguistics, Language Resources and Evaluation, {LREC/COLING} 2024,
                  20-25 May, 2024, Torino, Italy},
  pages        = {9615--9625},
  publisher    = {{ELRA} and {ICCL}},
  year         = {2024},
  url          = {https://aclanthology.org/2024.lrec-main.840},
  bibsource    = {dblp computer science bibliography, https://dblp.org}
}

@misc{katz2024backwardlensprojectinglanguage,
      title={Backward Lens: Projecting Language Model Gradients into the Vocabulary Space}, 
      author={Shahar Katz and Yonatan Belinkov and Mor Geva and Lior Wolf},
      year={2024},
      eprint={2402.12865},
      archivePrefix={arXiv},
      primaryClass={cs.CL},
      url={https://arxiv.org/abs/2402.12865}, 
}

@article{gurnee2026verbalizable,
  author={Gurnee, Wes and Sofroniew, Nicholas and Pearce, Adam and Piotrowski, Mateusz and Kauvar, Isaac and Chen, Runjin and Soligo, Anna and Bogdan, Paul and Ong, Euan and Wang, Rowan and Thompson, Ben and Abrahams, David and Kantamneni, Subhash and Ameisen, Emmanuel and Batson, Joshua and Lindsey, Jack},
  title={Verbalizable Representations Form a Global Workspace in Language Models},
  journal={Transformer Circuits Thread},
  year={2026},
  url={https://transformer-circuits.pub/2026/workspace/index.html}
}

@misc{geva2022transformerfeedforwardlayersbuild,
      title={Transformer Feed-Forward Layers Build Predictions by Promoting Concepts in the Vocabulary Space}, 
      author={Mor Geva and Avi Caciularu and Kevin Ro Wang and Yoav Goldberg},
      year={2022},
      eprint={2203.14680},
      archivePrefix={arXiv},
      primaryClass={cs.CL},
      url={https://arxiv.org/abs/2203.14680}, 
}

@inproceedings{merullo2023,
  author       = {Jack Merullo and
                  Louis Castricato and
                  Carsten Eickhoff and
                  Ellie Pavlick},
  title        = {Linearly Mapping from Image to Text Space},
  booktitle    = {The Eleventh International Conference on Learning Representations,
                  {ICLR} 2023, Kigali, Rwanda, May 1-5, 2023},
  publisher    = {OpenReview.net},
  year         = {2023},
  url          = {https://openreview.net/forum?id=8tYRqb05pVn},
  bibsource    = {dblp computer science bibliography, https://dblp.org}
}

@inproceedings{tsimpoukelli2021,
  author       = {Maria Tsimpoukelli and
                  Jacob Menick and
                  Serkan Cabi and
                  S. M. Ali Eslami and
                  Oriol Vinyals and
                  Felix Hill},
  editor       = {Marc'Aurelio Ranzato and
                  Alina Beygelzimer and
                  Yann N. Dauphin and
                  Percy Liang and
                  Jennifer Wortman Vaughan},
  title        = {Multimodal Few-Shot Learning with Frozen Language Models},
  booktitle    = {Advances in Neural Information Processing Systems 34: Annual Conference
                  on Neural Information Processing Systems 2021, NeurIPS 2021, December
                  6-14, 2021, virtual},
  pages        = {200--212},
  year         = {2021},
  url          = {https://proceedings.neurips.cc/paper/2021/hash/01b7575c38dac42f3cfb7d500438b875-Abstract.html},
  bibsource    = {dblp computer science bibliography, https://dblp.org}
}

@inproceedings{manas-etal-2023-mapl,
    title = "{MAPL}: Parameter-Efficient Adaptation of Unimodal Pre-Trained Models for Vision-Language Few-Shot Prompting",
    author = "Ma{\~n}as, Oscar  and
      Rodriguez Lopez, Pau  and
      Ahmadi, Saba  and
      Nematzadeh, Aida  and
      Goyal, Yash  and
      Agrawal, Aishwarya",
    editor = "Vlachos, Andreas  and
      Augenstein, Isabelle",
    booktitle = "Proceedings of the 17th Conference of the European Chapter of the Association for Computational Linguistics",
    month = may,
    year = "2023",
    address = "Dubrovnik, Croatia",
    publisher = "Association for Computational Linguistics",
    url = "https://aclanthology.org/2023.eacl-main.185/",
    doi = "10.18653/v1/2023.eacl-main.185",
    pages = "2523--2548"
}

@article{clipcap,
  author       = {Ron Mokady and
                  Amir Hertz and
                  Amit H. Bermano},
  title        = {ClipCap: {CLIP} Prefix for Image Captioning},
  journal      = {CoRR},
  volume       = {abs/2111.09734},
  year         = {2021},
  url          = {https://arxiv.org/abs/2111.09734},
  eprinttype   = {arXiv},
  eprint       = {2111.09734},
  bibsource    = {dblp computer science bibliography, https://dblp.org}
}

@misc{sourcemodality,
      title={Source-Modality Monitoring in Vision-Language Models}, 
      author={Etha Tianze Hua and Tian Yun and Ellie Pavlick},
      year={2026},
      eprint={2604.22038},
      archivePrefix={arXiv},
      primaryClass={cs.CL},
      url={https://arxiv.org/abs/2604.22038}, 
}

@article{papadimitriou2025,
  author       = {Isabel Papadimitriou and
                  Huangyuan Su and
                  Thomas Fel and
                  Naomi Saphra and
                  Sham M. Kakade and
                  Stephanie Gil},
  title        = {Interpreting the Linear Structure of Vision-language Model Embedding
                  Spaces},
  journal      = {CoRR},
  volume       = {abs/2504.11695},
  year         = {2025},
  url          = {https://doi.org/10.48550/arXiv.2504.11695},
  doi          = {10.48550/ARXIV.2504.11695},
  eprinttype   = {arXiv},
  eprint       = {2504.11695},
  bibsource    = {dblp computer science bibliography, https://dblp.org}
}

@article{platonic,
  author       = {Minyoung Huh and
                  Brian Cheung and
                  Tongzhou Wang and
                  Phillip Isola},
  title        = {The Platonic Representation Hypothesis},
  journal      = {CoRR},
  volume       = {abs/2405.07987},
  year         = {2024},
  url          = {https://doi.org/10.48550/arXiv.2405.07987},
  doi          = {10.48550/ARXIV.2405.07987},
  eprinttype   = {arXiv},
  eprint       = {2405.07987},
  bibsource    = {dblp computer science bibliography, https://dblp.org}
}

@article{DEHAENE2011254,
title = {The unique role of the visual word form area in reading},
journal = {Trends in Cognitive Sciences},
volume = {15},
number = {6},
pages = {254-262},
year = {2011},
issn = {1364-6613},
doi = {https://doi.org/10.1016/j.tics.2011.04.003},
url = {https://www.sciencedirect.com/science/article/pii/S1364661311000738},
author = {Stanislas Dehaene and Laurent Cohen}
}

@article{McCandliss2003,
  author = {McCandliss, Bruce D. and Cohen, Laurent and Dehaene, Stanislas},
  year = {2003},
  month = {7},
  title = {The visual word form area: expertise for reading in the fusiform gyrus},
  journal = {Trends in Cognitive Sciences},
  volume = {7},
  number = {7},
  pages = {293--299},
  doi = {10.1016/s1364-6613(03)00134-7},
  pmid = {12860187},
  issn = {1364-6613}
}

@misc{venhoff2025visualrepresentationsmaplanguage,
      title={How Visual Representations Map to Language Feature Space in Multimodal LLMs}, 
      author={Constantin Venhoff and Ashkan Khakzar and Sonia Joseph and Philip Torr and Neel Nanda},
      year={2025},
      eprint={2506.11976},
      archivePrefix={arXiv},
      primaryClass={cs.CV},
      url={https://arxiv.org/abs/2506.11976}, 
}

@misc{chen2023interpretingcontrollingvisionfoundation,
      title={Interpreting and Controlling Vision Foundation Models via Text Explanations}, 
      author={Haozhe Chen and Junfeng Yang and Carl Vondrick and Chengzhi Mao},
      year={2023},
      eprint={2310.10591},
      archivePrefix={arXiv},
      primaryClass={cs.CV},
      url={https://arxiv.org/abs/2310.10591}, 
}

@inproceedings{NEURIPS2024_ec3c79dc,
 author = {Shukor, Mustafa and Cord, Matthieu},
 booktitle = {Advances in Neural Information Processing Systems},
 doi = {10.52202/079017-4159},
 editor = {A. Globerson and L. Mackey and D. Belgrave and A. Fan and U. Paquet and J. Tomczak and C. Zhang},
 pages = {130848--130886},
 publisher = {Curran Associates, Inc.},
 title = {Implicit Multimodal Alignment: On the Generalization of Frozen LLMs to Multimodal Inputs},
 url = {https://proceedings.neurips.cc/paper_files/paper/2024/file/ec3c79dc0c2b85532cfd1012a4aaa923-Paper-Conference.pdf},
 volume = {37},
 year = {2024}
}

@inproceedings{schwettmann2023,
  author       = {Sarah Schwettmann and
                  Neil Chowdhury and
                  Samuel Klein and
                  David Bau and
                  Antonio Torralba},
  title        = {Multimodal Neurons in Pretrained Text-Only Transformers},
  booktitle    = {{IEEE/CVF} International Conference on Computer Vision, {ICCV} 2023
                  - Workshops, Paris, France, October 2-6, 2023},
  pages        = {2854--2859},
  publisher    = {{IEEE}},
  year         = {2023},
  url          = {https://doi.org/10.1109/ICCVW60793.2023.00308},
  doi          = {10.1109/ICCVW60793.2023.00308},
  bibsource    = {dblp computer science bibliography, https://dblp.org}
}

@inproceedings{venhoff2025toolatetorecall,
  author       = {Constantin Venhoff and
                  Ashkan Khakzar and
                  Sonia Joseph and
                  Philip H. S. Torr and
                  Neel Nanda},
  editor       = {Danielle Belgrave and
                  Cheng Zhang and
                  Laura N. Montoya and
                  Hsuan{-}Tien Lin and
                  Razvan Pascanu and
                  Piotr Koniusz and
                  Marzyeh Ghassemi and
                  Nancy Chen and
                  Iv{\'{a}}n Vladimir Meza Ru{\'{\i}}z and
                  Arturo Loaiza{-}Bonilla},
  title        = {Too Late to Recall: Explaining the Two-Hop Problem in Multimodal Knowledge
                  Retrieval},
  booktitle    = {Advances in Neural Information Processing Systems 38: Annual Conference
                  on Neural Information Processing Systems 2025, NeurIPS 2025, San Diego,
                  CA, USA, December 2-7, 2025 / Mexico City, Mexico, November 30 - December
                  5, 2025},
  year         = {2025},
  url          = {http://papers.nips.cc/paper\_files/paper/2025/hash/4526cfacdbca6b6e184568dac91bf070-Abstract-Conference.html},
  bibsource    = {dblp computer science bibliography, https://dblp.org}
}

@misc{karamolegkou2026readingguessingvisualgrounding,
      title={Reading or Guessing? Visual Grounding Failures of Vision-Language Models for OCR in Ancient Greek Editions}, 
      author={Antonia Karamolegkou and Nicolas Angleraud and Benoît Sagot and Thibault Clérice},
      year={2026},
      eprint={2605.27750},
      archivePrefix={arXiv},
      primaryClass={cs.CL},
      url={https://arxiv.org/abs/2605.27750}, 
}

@misc{liang2026visualmeritlinguisticcrutch,
      title={Visual Merit or Linguistic Crutch? A Close Look at DeepSeek-OCR}, 
      author={Yunhao Liang and Ruixuan Ying and Bo Li and Hong Li and Kai Yan and Qingwen Li and Min Yang and Okamoto Satoshi and Zhe Cui and Shiwen Ni},
      year={2026},
      eprint={2601.03714},
      archivePrefix={arXiv},
      primaryClass={cs.CL},
      url={https://arxiv.org/abs/2601.03714}, 
}

@misc{he2025seeingbelievingmitigatingocr,
      title={Seeing is Believing? Mitigating OCR Hallucinations in Multimodal Large Language Models}, 
      author={Zhentao He and Can Zhang and Ziheng Wu and Zhenghao Chen and Yufei Zhan and Yifan Li and Zhao Zhang and Xian Wang and Minghui Qiu},
      year={2025},
      eprint={2506.20168},
      archivePrefix={arXiv},
      primaryClass={cs.CV},
      url={https://arxiv.org/abs/2506.20168}, 
}

@misc{shu2025semanticsmisleadvisionmitigating,
      title={When Semantics Mislead Vision: Mitigating Large Multimodal Models Hallucinations in Scene Text Spotting and Understanding}, 
      author={Yan Shu and Hangui Lin and Yexin Liu and Yan Zhang and Gangyan Zeng and Yan Li and Yu Zhou and Ser-Nam Lim and Harry Yang and Nicu Sebe},
      year={2025},
      eprint={2506.05551},
      archivePrefix={arXiv},
      primaryClass={cs.CV},
      url={https://arxiv.org/abs/2506.05551}, 
}

@article{ethaconflicting2025,
  author       = {Tianze Hua and
                  Tian Yun and
                  Ellie Pavlick},
  title        = {How Do Vision-Language Models Process Conflicting Information Across
                  Modalities?},
  journal      = {CoRR},
  volume       = {abs/2507.01790},
  year         = {2025},
  url          = {https://doi.org/10.48550/arXiv.2507.01790},
  doi          = {10.48550/ARXIV.2507.01790},
  eprinttype   = {arXiv},
  eprint       = {2507.01790},
  bibsource    = {dblp computer science bibliography, https://dblp.org}
}

@article{ioi,
  author       = {Kevin Wang and
                  Alexandre Variengien and
                  Arthur Conmy and
                  Buck Shlegeris and
                  Jacob Steinhardt},
  title        = {Interpretability in the Wild: a Circuit for Indirect Object Identification
                  in {GPT-2} small},
  journal      = {CoRR},
  volume       = {abs/2211.00593},
  year         = {2022},
  url          = {https://doi.org/10.48550/arXiv.2211.00593},
  doi          = {10.48550/ARXIV.2211.00593},
  eprinttype   = {arXiv},
  eprint       = {2211.00593},
  bibsource    = {dblp computer science bibliography, https://dblp.org}
}

@article{vlmsneedwords,
  author       = {Haz Sameen Shahgir and
                  Xiaofu Chen and
                  Yu Fu and
                  Erfan Shayegani and
                  Nael B. Abu{-}Ghazaleh and
                  Yova Kementchedjhieva and
                  Yue Dong},
  title        = {VLMs Need Words: Vision Language Models Ignore Visual Detail In Favor
                  of Semantic Anchors},
  journal      = {CoRR},
  volume       = {abs/2604.02486},
  year         = {2026},
  url          = {https://doi.org/10.48550/arXiv.2604.02486},
  doi          = {10.48550/ARXIV.2604.02486},
  eprinttype   = {arXiv},
  eprint       = {2604.02486},
  bibsource    = {dblp computer science bibliography, https://dblp.org}
}

@misc{tartaglini2025bottlenecks,
      title={Diagnosing Bottlenecks in Data Visualization Understanding by Vision-Language Models}, 
      author={Alexa R. Tartaglini and Satchel Grant and Daniel Wurgaft and Christopher Potts and Judith E. Fan},
      year={2025},
      eprint={2510.21740},
      archivePrefix={arXiv},
      primaryClass={cs.CV},
      url={https://arxiv.org/abs/2510.21740}, 
}

@misc{paraselli2026slowseeslowsuppress,
      title={Slow to See, Slow to Suppress: Understanding the Effects of Modality in Context-Memory Conflicts}, 
      author={Athulith Paraselli and Etha Tianze Hua and Ellie Pavlick},
      year={2026},
      eprint={2609.00293},
      archivePrefix={arXiv},
      primaryClass={cs.CL},
      url={https://arxiv.org/abs/2609.00293}, 
}

@inproceedings{aurora,
  author       = {Benno Krojer and
                  Dheeraj Vattikonda and
                  Luis Lara and
                  Varun Jampani and
                  Eva Portelance and
                  Chris Pal and
                  Siva Reddy},
  editor       = {Amir Globersons and
                  Lester Mackey and
                  Danielle Belgrave and
                  Angela Fan and
                  Ulrich Paquet and
                  Jakub M. Tomczak and
                  Cheng Zhang},
  title        = {Learning Action and Reasoning-Centric Image Editing from Videos and
                  Simulation},
  booktitle    = {Advances in Neural Information Processing Systems 37: Annual Conference
                  on Neural Information Processing Systems 2024, NeurIPS 2024, Vancouver,
                  BC, Canada, December 10 - 15, 2024},
  year         = {2024},
  url          = {http://papers.nips.cc/paper\_files/paper/2024/hash/434d512d6d79a506fd32f8b39abb7c19-Abstract-Datasets\_and\_Benchmarks\_Track.html},
  bibsource    = {dblp computer science bibliography, https://dblp.org}
}

@book{Weinberger2016,
  author = {Weinberger, Eliot},
  title = {19 ways of looking at Wang Wei: with more ways},
  publisher = {New Directions Books},
  address = {New York, NY},
  year = {2016},
  pages = {88},
  note = {Afterword by Octavio Paz}
}

@inproceedings{Chang2025,
  author = {Chang, Adrian and Feucht, Sheridan and Wallace, Byron C and Bau, David},
  title = {Does {FLUX} Know What It's Writing?},
  booktitle = {NeurIPS Mechanistic Interpretability Workshop},
  year = {2025},
  address = {San Diego}
}

@inproceedings{baek,
    title = "How Do Large Vision-Language Models See Text in Image? Unveiling the Distinctive Role of {OCR} Heads",
    author = "Baek, Ingeol  and
      Chang, Hwan  and
      Ryu, Sunghyun  and
      Lee, Hwanhee",
    editor = "Christodoulopoulos, Christos  and
      Chakraborty, Tanmoy  and
      Rose, Carolyn  and
      Peng, Violet",
    booktitle = "Proceedings of the 2025 Conference on Empirical Methods in Natural Language Processing",
    month = nov,
    year = "2025",
    address = "Suzhou, China",
    publisher = "Association for Computational Linguistics",
    url = "https://aclanthology.org/2025.emnlp-main.1032/",
    doi = "10.18653/v1/2025.emnlp-main.1032",
    pages = "20441--20453",
    ISBN = "979-8-89176-332-6"
}
}

\newpage
\clearpage
\appendix

\section{Model Details}

In this work, we study four models of varying sizes with three distinct model families. See Table~\ref{tab:models} for architectural details of models studied in this paper. 

\begin{table*}[h]
\caption{Architectural details of models used in this paper. ``Tied'' refers to tied embeddings, ``Img Encoder'' refers to whether the image encoder is trained end-to-end or frozen, and ``Single-tok COCO'' is the number of COCO \cite{COCO} categories that are single-token for this tokenizer (which determines results in Sections~\ref{sec:confidence}-\ref{sec:localization}).}\label{tab:models}
\centering
\begin{tabular}{@{}llllclc@{}}
\toprule
Full Model Name    & Cite  & Abbrv. & Total Heads & Tied? & Img Encoder? & Single-tok COCO \\ \midrule
\small Qwen/Qwen3-VL-2B-Instruct & \cite{bai2025qwen3vltechnicalreport}  & \small Qwen3-VL-2B & \small 28 lyrs $\times$ 16 hds = 448           & Yes &  End-to-end & 56/80   \\
\small Qwen/Qwen3-VL-8B-Instruct & \cite{bai2025qwen3vltechnicalreport}   & Qwen3-VL-8B & \small 36 lyrs $\times$ 32 hds = 1152         & No & End-to-end & 56/80    \\ 
\small allenai/Molmo2-O-7B & \cite{molmo2} & \small Molmo2-7B & \small 32 lyrs $\times$ 32 hds = 1024 & No & End-to-end & 56/80 \\
\small llava-hf/llava-v1.6-34b-hf & \cite{liu2024llavanext} & \small Llava-Next-34B & \small 60 lyrs $\times$ 56 hds = 3360 & No & End-to-end & 51/80 \\
\bottomrule
\end{tabular}
\end{table*}

\begin{table}[h]
\caption{OCR Accuracy for randomly-selected English words superimposed on either ImageNet or white backgrounds ($n=100$). Accuracies are obtained with the prompt ``Transcribe the word in this image.'' and assistant prefill ``Word:''.}\label{tab:ocr_acc}
\centering
\begin{tabular}{@{}lcc@{}}
\toprule
Model      & ImageNet Bg & White Bg \\ \midrule
Qwen3-VL-2B  & 0.89 & 0.07                 \\
Qwen3-VL-8B  & 0.93 & 0.05               \\ 
Molmo2-O-7B  & 0.90 & 0.44 \\
Llava-Next-34B & 0.80 & 0.75 \\
\bottomrule
\end{tabular}
\end{table}

\section{Finding Verbalization Heads}\label{app:score-other-models}

\begin{figure}[h]
\centering
\includegraphics[width=0.15\textwidth]{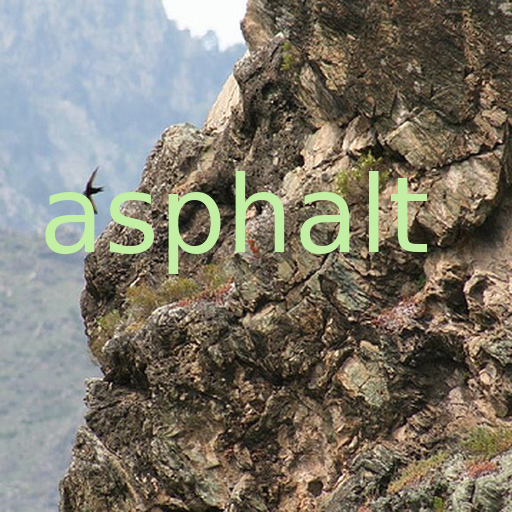}
\includegraphics[width=0.15\textwidth]{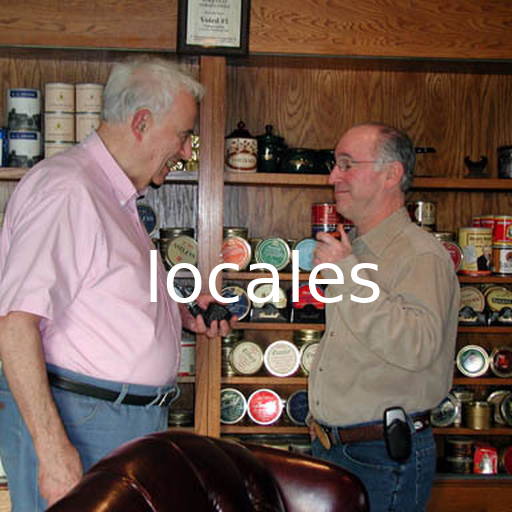}
\includegraphics[width=0.15\textwidth]{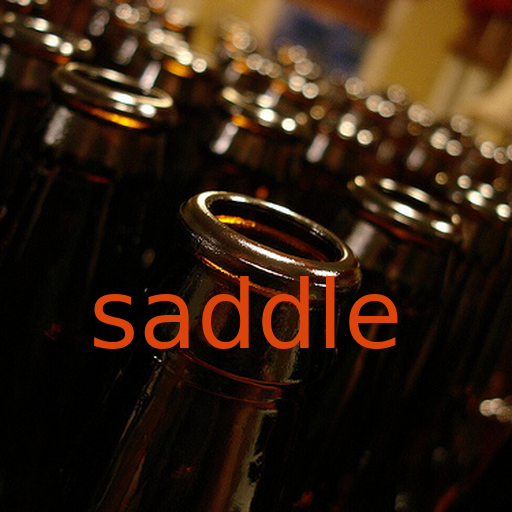}
\caption{Random examples from our OCR dataset used to identify heads in Section~\ref{sec:finding-heads}.}
\label{fig:ocr_dataset_examples}
\end{figure}

OCR dataset examples are shown in Figure~\ref{fig:ocr_dataset_examples}. We use large font sizes and randomly sample colors that contrast with the background. Table~\ref{tab:ocr_acc} shows model accuracy for these stimuli. Due to models having poor OCR performance on stimuli with white backgrounds, we opt for realistic backgrounds.

\begin{figure}[h]
    \centering
    \includegraphics[width=\linewidth]{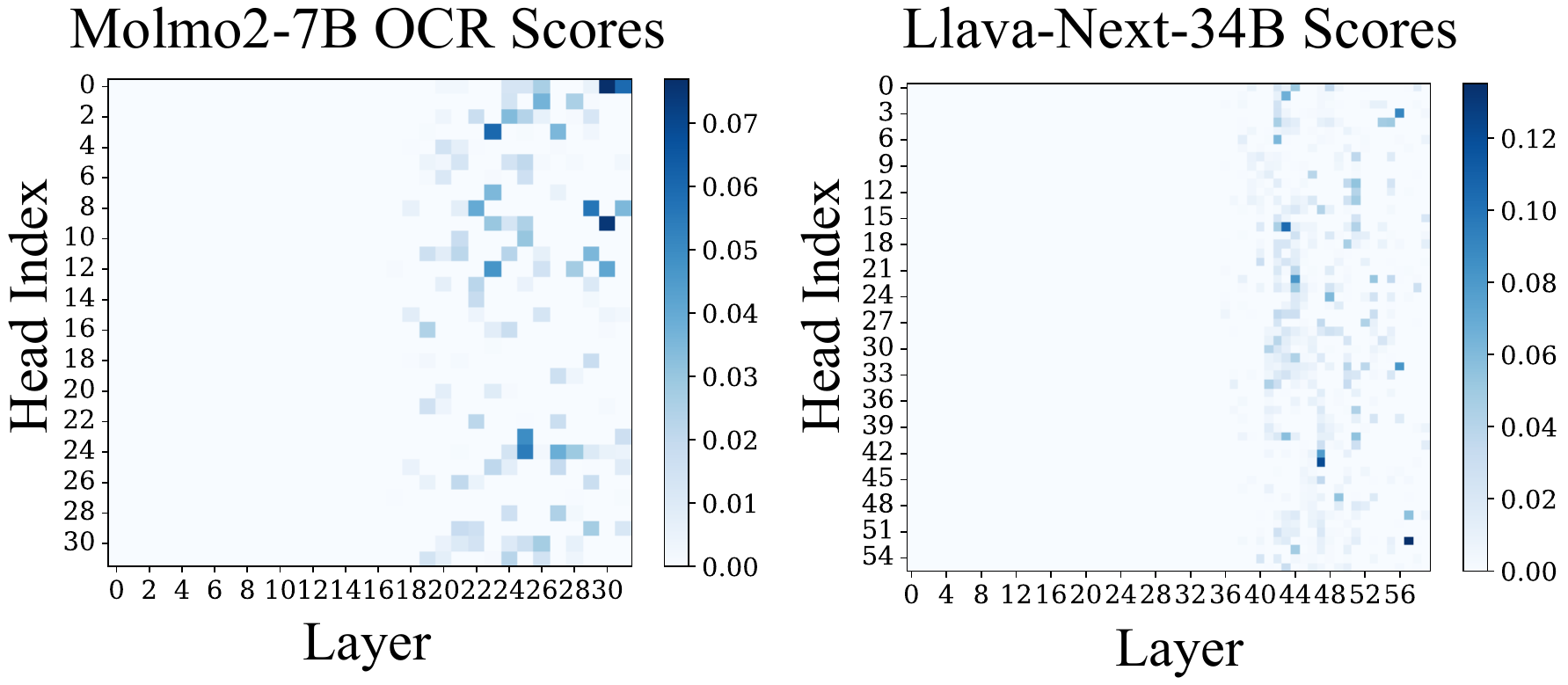}
    \caption{OCR scores from Equation~\ref{eq:ocr_score} for Molmo2-O-7B and Llava-Next-34B. The takeaway is similar to Figure~\ref{fig:qwen-heatmaps} for Qwen3-VL models: OCR heads appear in late layers.}
    \label{fig:additional-heatmaps}
\end{figure}

\begin{figure}[h]
    \centering
    \includegraphics[width=\linewidth]{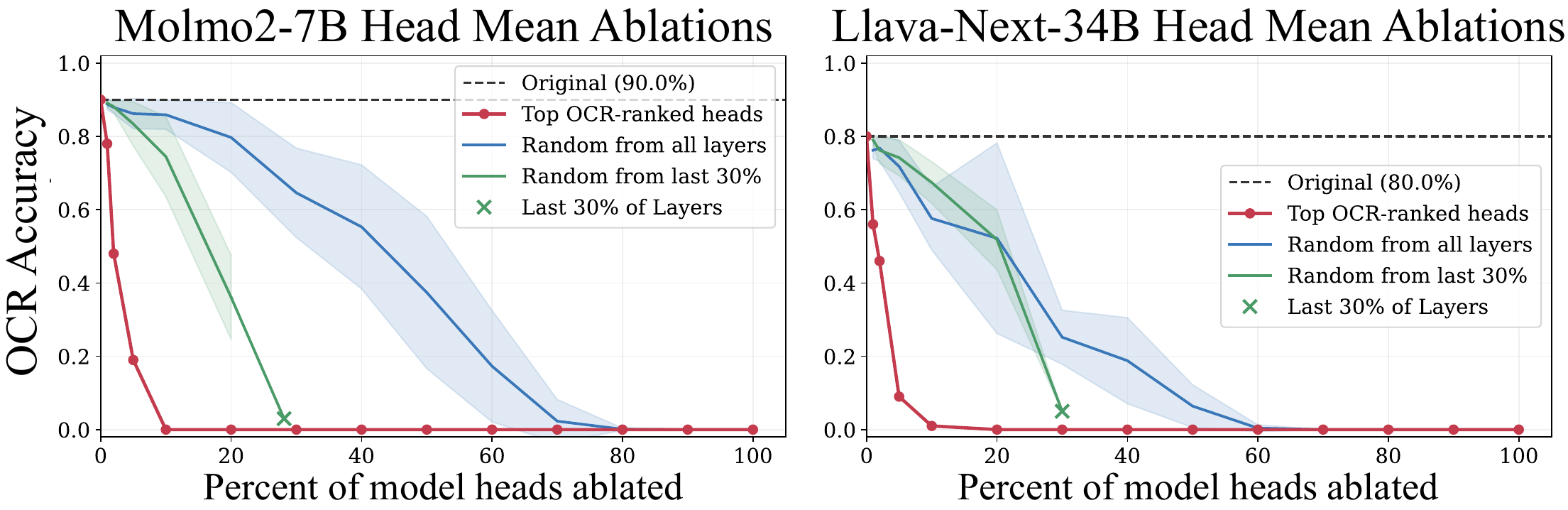}
    \caption{Progressively mean-ablating top-scoring OCR heads for Molmo2-O-7B and Llava-Next-34B. Results are similar to Qwen3-VL results from Figure~\ref{fig:head-ablations}.}
    \label{fig:other-model-ablations}
\end{figure}

\begin{figure*}[h]
    \centering
    \includegraphics[width=\linewidth]{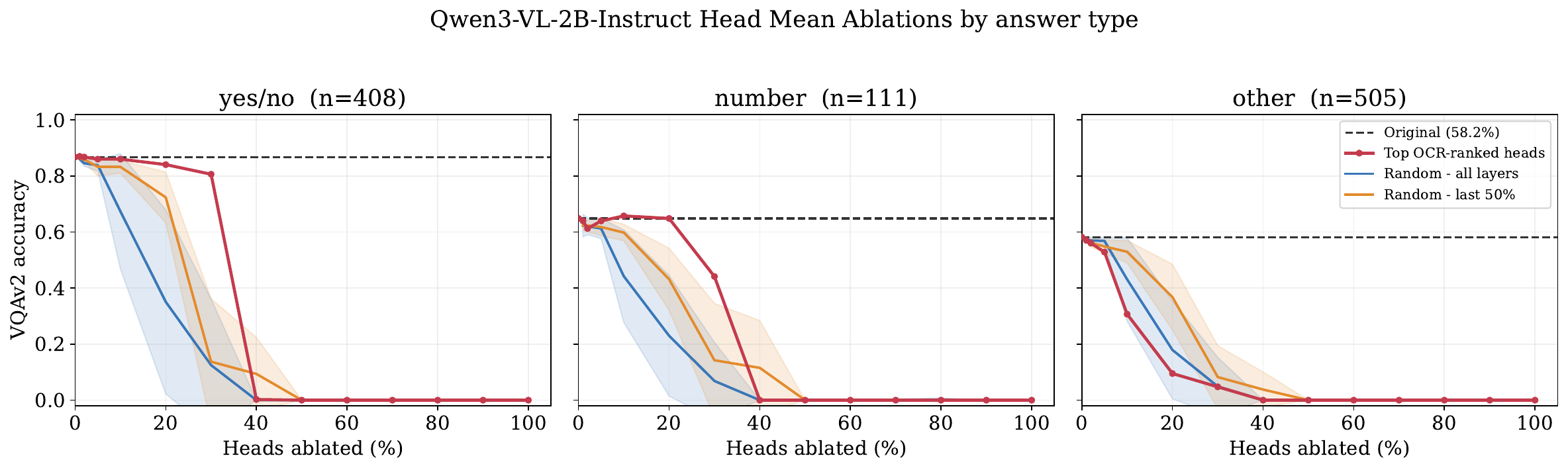}
    \caption{We also measure accuracy for VQAv2 \cite{balanced_vqa_v2} when progressively mean-ablating the top-scoring OCR heads throughout generation. While ablation of these heads does not affect performance for questions that have Yes/No or numeric answers, it does affect performance for other types of questions (e.g., ``What is the type of food in this image?'' $\rightarrow$ ``Mexican'') slightly more than an equal number of random heads. This suggests that verbalization heads may be useful for more general-purpose question answering tasks.}
    \label{fig:vqa-ablations}
\end{figure*}

Results from Section~\ref{sec:finding-heads} are shown for Molmo2-7B and Llava-Next-34B in Figure~\ref{fig:additional-heatmaps} and Figure~\ref{fig:other-model-ablations}. We also progressively ablate heads responsible for OCR for Qwen3-VL-2B on VQA in Figure~\ref{fig:vqa-ablations}.

\section{Overlap with Other Heads}\label{app:overlap}
Attention patterns from Figure~\ref{fig:switching-attn} are reminiscent of \emph{filter heads} from prior work \cite{sensharma2023filter}. Are they the same heads? We find that of the 11 filter heads found in Qwen3-1.7B \cite{yang2025qwen3technicalreport} (the base LLM for Qwen3-VL-2B-Instruct), six of them are also in the top-10\% of verbalization heads. This suggests that vision fine-tuning may cause filter heads to ``evolve'' into verbalization heads.

We also check for overlap with \emph{gaze heads} \cite{gandikota2026gazeheads}, a type of attention head in VLMs that keeps track of which portion of an image the VLM is captioning. For Qwen3-VL-2B, 30\% of gaze heads are also verbalization heads (3/10), and for Qwen3-VL-8B, 29\% of gaze heads are also verbalization heads (29/100). This is higher than the expected overlap when sampling the same number of random heads.\footnote{For Qwen3-VL-2B, we have 45 verbalization heads and 10 gaze heads, so the expected number of overlapping heads is (10$\times$45)/448, or 1/10 gaze heads. For Qwen3-VL-8B, we have 115 verbalization heads and 100 gaze heads, so the expected overlap with random samples is (100$\times$115)/1152, or 10/100 gaze heads.}
Although this means that verbalization heads are somewhat distinct from gaze heads, the modest overlap also supports our claim that analysis of OCR can identify heads more generally responsible for mapping from pixels to semantics. 

\section{Full Verbalization Lens Examples}\label{app:lens_qual_allmodels}

In Figures~\ref{fig:pathway_qwen2b}, \ref{fig:pathway_molmo2}, and \ref{fig:pathway_llava}, we show full verbalization lens outputs for models not shown in Figure~\ref{fig:lens_qual}. All outputs are for layer 0, and compared to raw logit lens outputs at layer 0. 




\section{Object Confidence}\label{app:roc}

For each category, we sample $n$=512 images containing $o$ and $n$ images without $o$, except for Llava-Next-34B, for which $n$=128 due to computational constraints. Figure~\ref{fig:roc} shows these results using ROC-AUC curves, instead of calculating average difference in maximum $P(o)$ as in the main paper. Interestingly, ROC-AUC shows that raw logit lens has more signal for Qwen models, despite verbalization lens still showing marked improvements.


\section{Object Localization Details}\label{app:localization}

Here, we provide implementation details for Section~\ref{sec:localization}. 
For each image, we score one segmentation map per object category
present in that image, where the ground truth mask is the union of all instances of that category (e.g., if there are multiple people in an image, our ground truth mask includes all of those people).
For the 512 images that we display numbers for in Table~\ref{tab:localization-allmodels}, there are 1216 (image, category) pairs, an average of just over two object categories per image. Averaging over these pairs ensures that each object counts once regardless of image/object size. 

We pick a threshold $\tau$ for mIoU based on a disjoint set of 128 images. We show chosen mIoU thresholds in Table~\ref{tab:taus}.
We omit token-level accuracy, as in our dataset, only 15.7\% of tokens are positive. Therefore, a baseline that never segments any objects can trivially achieve 84.3\% token accuracy. 

\begin{table}[]
    \centering
    \caption{Thresholds $\tau$ for each approach in Table~\ref{tab:localization-allmodels}, chosen based on a disjoint set of 128 images. We sweep over a log-spaced range of 60 thresholds based on the range of probability scores for that method.}
    \begin{tabular}{lcc}
        \toprule
        \multicolumn{1}{c}{} & \multicolumn{1}{c}{\textbf{Qwen3-VL-8B}} & \multicolumn{1}{c}{\textbf{Qwen3-VL-2B}} \\
        \midrule
        Random Heads & $7.2{\times}10^{-6}$ & $5.1{\times}10^{-8}$ \\
        Raw Logit Lens & $3.0{\times}10^{-4}$ & $4.4{\times}10^{-3}$ \\
        Verbalization Lens & $3.4{\times}10^{-3}$ & $1.1{\times}10^{-3}$ \\
        \bottomrule
    \end{tabular}
    \label{tab:taus}
\end{table}

\section{LatentLens Comparison Details}\label{app:latentlens-details}

We describe our setup for the LatentLens \citep{latentlens} comparison from Section~\ref{sec:comparing-latentlens} in more detail here.

\paragraph{LatentLens replication.} For each of the four models, we build a LatentLens contextual index following \citet{latentlens}'s released implementation and corpus. 
We extract contextual embeddings at every model's embedding layer plus a set of decoder layers space across the model, and store up to 30 contexts per unique token. Following \cite{latentlens}, we retrieve the top-$k$ nearest neighbors by cosine similarity across all indexed layers jointly at query time. As in their evaluation, LatentLens neighbor tokens are often subwords; we expand each to its full containing word using the neighbor's own source sentence as context (e.g.\ ``ing'' $\to$ ``rendering''). 

We find that a small fraction of nearest-neighbor matches (concentrated in Qwen3-VL-8B's middle layers) are whitespace tokens carrying no semantic content despite high cosine similarity. Thus, we search $k=32$ neighbors and take the top-5 non-blank words.

\paragraph{Layers.} For each model we evaluate six layers: Qwen3-VL-2B $\{0, 8, 12, 16, 20, 27\}$, Qwen3-VL-8B $\{0, 8, 12, 20, 24, 35\}$, Molmo2-O-7B $\{0, 8, 12, 19, 22, 31\}$, Llava-Next-34B $\{0, 10, 24, 36, 42, 59\}$. See Table~\ref{tab:models} for model details.

\paragraph{Images.} We sample 100 random images (fixed seed) from the same COCO val2014 pool used elsewhere in the paper. For each image we decode one patch, chosen pseudorandomly (seeded per image) from the central 60\% of that model's patch grid to avoid picking pure background/padding; the same patch is used across all 6 layers for a given image. 

\paragraph{LLM judge.} We use the same LLM-judge as \citet{latentlens}. We show \texttt{GPT-5} the full image with a red bounding box over the target patch, a cropped close-up of that region, and the top-5 candidate words for one lens at a particular layer. The judge labels each candidate as \emph{concrete} (literally visible in the box), \emph{abstract} (an implied concept/activity/quality), or \emph{globally} related (visible elsewhere in the image but not the box). A patch is scored interpretable if at least one candidate word receives any of these three labels. 

\paragraph{Results.} Figure~\ref{fig:lens-judge-layers} shows the resulting interpretable-token percentage across layers for all four models. For all models, verbalization lens is on par with or better than LatentLens in terms of LLM judge scores. We note that verbalization lens is especially helpful for Llava-Next-34B. 


\section{LLM Judge for Editing}
\label{app:appendix-llm-judge}

Table~\ref{tab:llm-judge-questions} lists LLM judge prompts for evaluating image edits from Section~\ref{sec:edit_eval}. For each question, the judge returns a structured response containing an integer \texttt{answer} (its rating).

\begin{table*}[h]
\centering
\small
\begin{tabular}{@{}p{2.3cm}p{10.5cm}@{}}
\toprule
\textbf{Question} & \textbf{Prompt template} \\
\midrule
Prevalence of $c$ &
On a scale from 0--10, how prominently does the concept of `\{$c$\}'
feature in this image caption? Let's say that a score of 10 means that the concept is the
main focus, 0 means that the concept is not mentioned at all, and in-between scores mean
that the concept is mentioned to some extent. If the caption is complete nonsense, rate it
as 0.\newline\newline
Caption:\textbackslash n\textbackslash n\texttt{\{caption\}} \\
\addlinespace
Specificity &
You will be shown two captions. The first is a Ground Truth caption describing a real
image. The second is a Counterfactual version of the first caption, where the concept of
`$c_{\text{rem}}$' has been replaced by the concept of
`$c_{\text{add}}$'. Please rate how similar the Counterfactual caption is to
the Ground Truth caption on a scale from 0--10. If the two captions are the exact same,
give a rating of 10; close paraphrases can also get high ratings. You can still give a high
score if the Counterfactual caption appears to describe a similar scene to the Ground Truth
caption when you ignore the edited concepts: for example, if we removed the concept
``motorcycle'' and replaced it with ``dog,'' then the captions ``a motorcycle in an empty
garage, dimly lit'' and ``a dog in a low-lit empty garage'' would get a score of 9, because
they are close paraphrases of each other when you ignore the edited concepts. In the worst
case, if the Counterfactual Caption has nothing to do with the Ground Truth Caption at all,
give a rating of 0.\newline\newline
Ground Truth Caption:\textbackslash n\textbackslash n\texttt{\{clean\_caption\}}\newline\newline
Counterfactual Caption:\textbackslash n\textbackslash n
\texttt{\{edited\_caption\}} \\
Coherence &
On a scale from 0--10, ignoring the semantics and focusing on grammar and writing quality,
how coherent is this image caption (with 10 being perfectly coherent and 0 being complete
gibberish)?\newline\newline
Caption:\textbackslash n\textbackslash n\texttt{\{caption\}} \\
\addlinespace
\bottomrule
\end{tabular}
\caption{Question templates for LLM judging from Section~\ref{sec:edit_eval}. We use the ``Prevalence of $c$'' template both for $c_{\text{rem}}$ and $c_{\text{add}}$. For Specificity scores, we also give the judge the model's original \texttt{clean\_caption}, so that it can evaluate whether any important details have changed.}
\label{tab:llm-judge-questions}
\end{table*}

\section{Editing Sweep for Rank/Scaling Factors}\label{app:edit_sweep}

We run an initial sweep over ranks and scaling factors for the edit in Section~\ref{sec:editing}. For each setting, we greedy-decode a caption per image. 
We choose to sweep across scaling factors $\alpha\in\{1, 2, 3, 4, 5, 10\}$. 

\begin{figure}
    \centering
    \includegraphics[width=\linewidth]{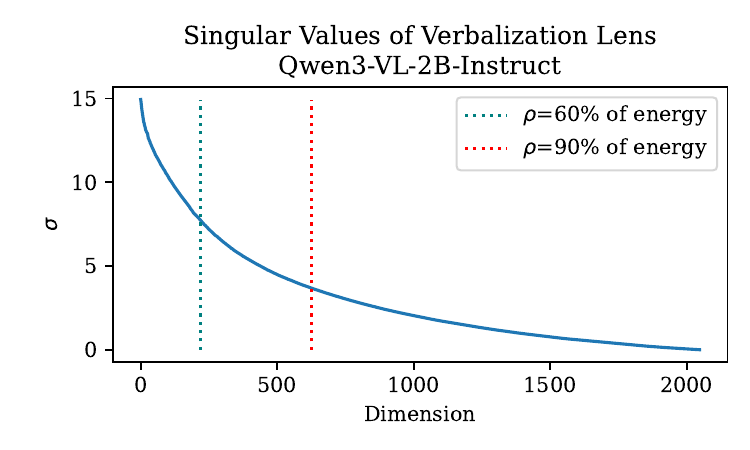}
    \caption{Singular values of $\mathbf{L}_p$ for Qwen3-VL-2B. We plot dotted lines to indicate the ranks that explain 60\% and 90\% of energy. See Section~\ref{sec:editing} for discussion of low-rank approximations of $\mathbf{L}_p$. }
    \label{fig:svs_qwen2b}
\end{figure}

Ranks are chosen for each model based on the number of dimensions it takes to explain $\rho$\% of the energy of $\mathbf{L}_p$. That is, if $\Sigma$=$[\sigma_1,...\sigma_{d_{\text{model}}}]$, we take the minimum top-$k$ dimensions such that $\text{sum}(\Sigma[:k]^2)/\text{sum}(\Sigma^2)>\rho/100$ for $\rho\in\{1, 5, 10, 20, 40, 60, 80, 90, 100\}$. For reference, we show the singular values of $\mathbf{L}_p$ for Qwen3-VL-2B in Figure~\ref{fig:svs_qwen2b}. 

Figures~\ref{fig:edit_sweep_2b} and \ref{fig:edit_sweep_8b} show results for LLM judge metrics from Section~\ref{sec:edit_eval} across $n$=100 randomly-sampled ImageNet images/object pairs \cite{imagenet}. We find that our edit generally requires a scaling factor $\alpha>1$, and does not have an effect for full-rank $\mathbf{L}^{-1}_p$. We also find that a large scaling factor $\alpha$=10 is damaging, as it causes specificity to drop. 

Thus, we choose $\rho$=60\% of verbalization lens energy and perform edits for $n$=256 additional images across scaling factors (excluding $\alpha$=10) in Figure~\ref{fig:edit_main}b. $\rho$ is constant for all qualitative generations shown in the paper. 

\section{Behind the Scenes}

Following \cite{aurora} and \cite{latentlens}, we describe the trajectory of this project in hopes of humanizing this work and making the research process transparent. 

This project began because SF was interested in writing and letters. They had worked on letter representations in text-to-image diffusion models before \cite{Chang2025}, but had just returned from an internship and were trying to decide how exactly to continue this line of work. With encouragement from Si Wu and John Cayley, they began working on a project (which is still in-progress) on VLM representations of Chinese poetry; specifically, the poem \chinese{鹿柴} by Wang Wei, which has an interesting history of analysis in terms of its translations \cite{Weinberger2016}.

In trying to understand how VLMs might process an \textit{image} of this poem, SF set out to first figure out how VLMs do OCR in general. The plan was just to figure out how OCR worked, so that this knowledge could be leveraged to create interesting visualizations for this art project. But then, SF realized that the heads they had picked out for OCR were much more general when they were playing around with the OV lens approach used in prior work \cite{feucht2025dualroute}. Clicking on the word ``Hey'' in a test image yielded that word, as expected. But out of curiosity they clicked on an emoji in the image, expecting the result to be garbage. Surprisingly, they saw the token \texttt{\_emoji} in early layers! In that moment, the authors realized that they had a pretty interesting little finding that could provide a nice approach to ``logit lens for image tokens.'' 

At the same time, HA and SW had been helping out SF with other lines of attack on understanding text in images (studying emoji and lexical representations). With this surprise, the authors decided to band together and write a paper on this finding. They decided to submit to a conference in a location SF wanted to visit for personal reasons. They recruited BK, who was just beginning his postdoc at the lab, because of his expertise from prior work on this topic, and started writing this paper. Even though BK was in the middle of moving in, and SF came down with a fever in the last few weeks of work on this paper, it ended up coming together in the end.

\begin{figure}
    \centering
    \includegraphics[width=\linewidth]{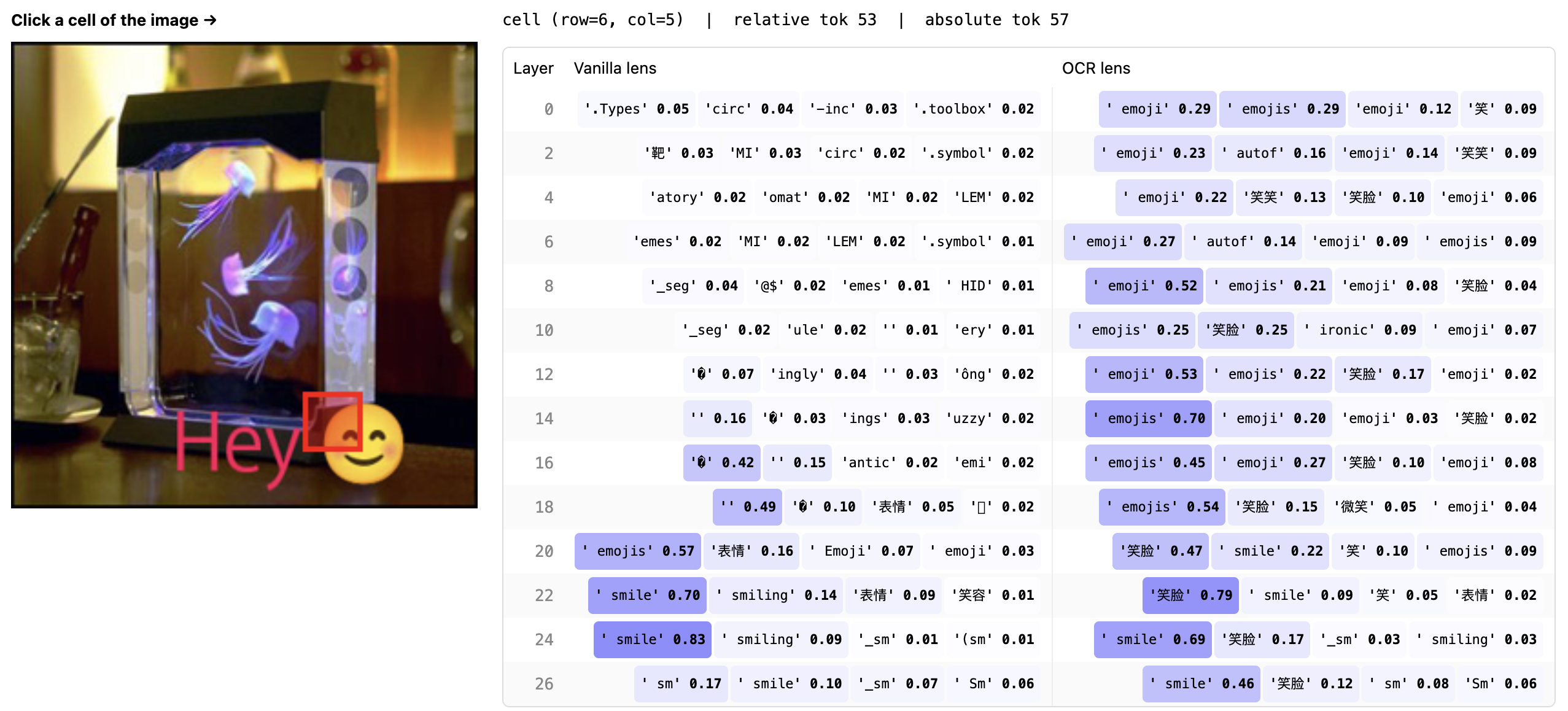}
    \caption{The original image that SF was using, which was just a random image with the text ``Hey'' and an emoji overlaid using Photoshop. Clicking on the emoji yielded \texttt{\_emoji}, even though SF expected to just see noise.}
    \label{fig:hey_jellyfish}
\end{figure}

\begin{figure*}[h]
    \centering
    \includegraphics[width=0.75\linewidth]{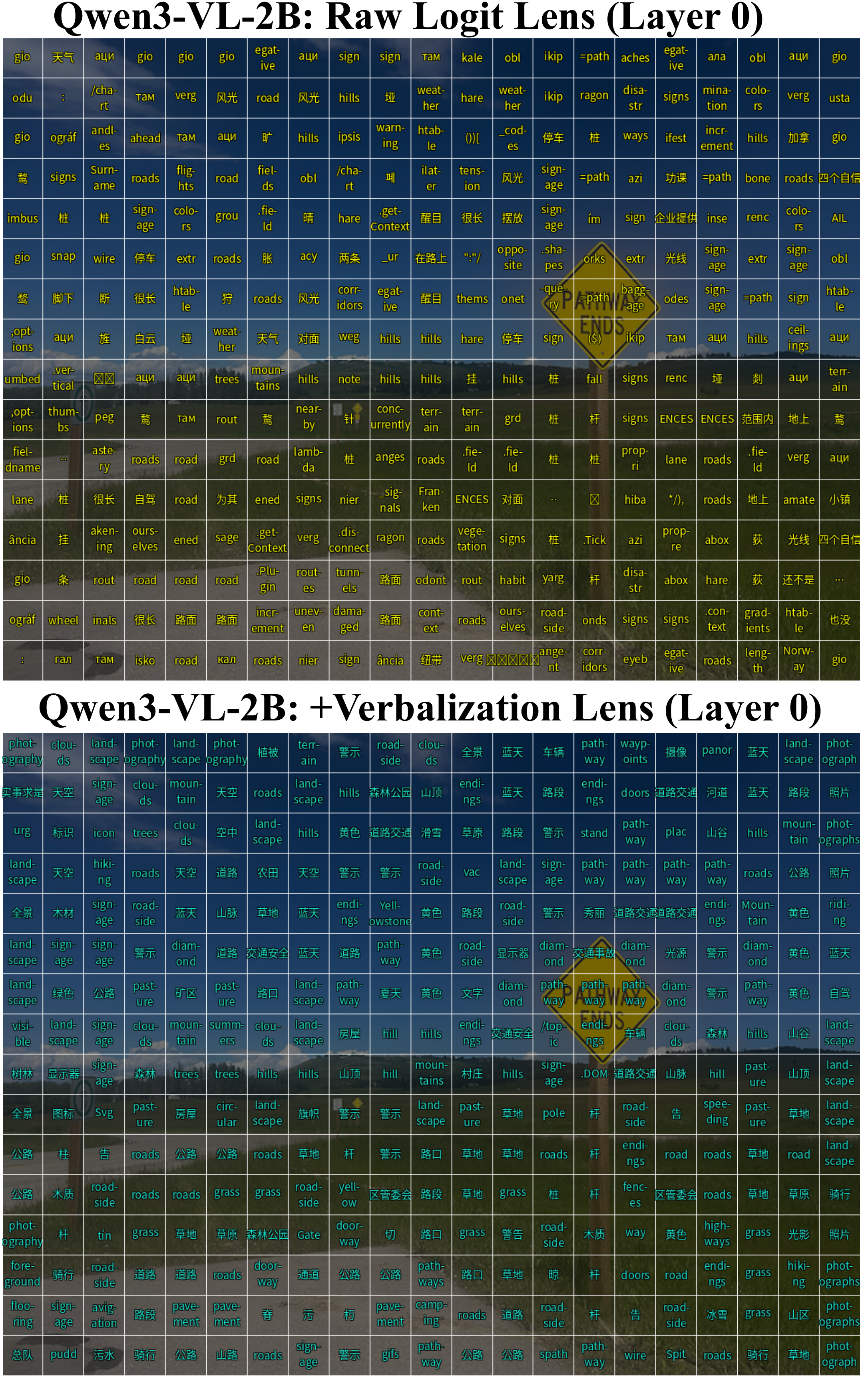}
    \caption{Verbalization lens results for image in Figure~\ref{fig:lens_qual} for \textbf{Qwen3-VL-2B}. }
    \label{fig:pathway_qwen2b}
\end{figure*}

\begin{figure*}[h]
    \centering
    \includegraphics[width=0.75\linewidth]{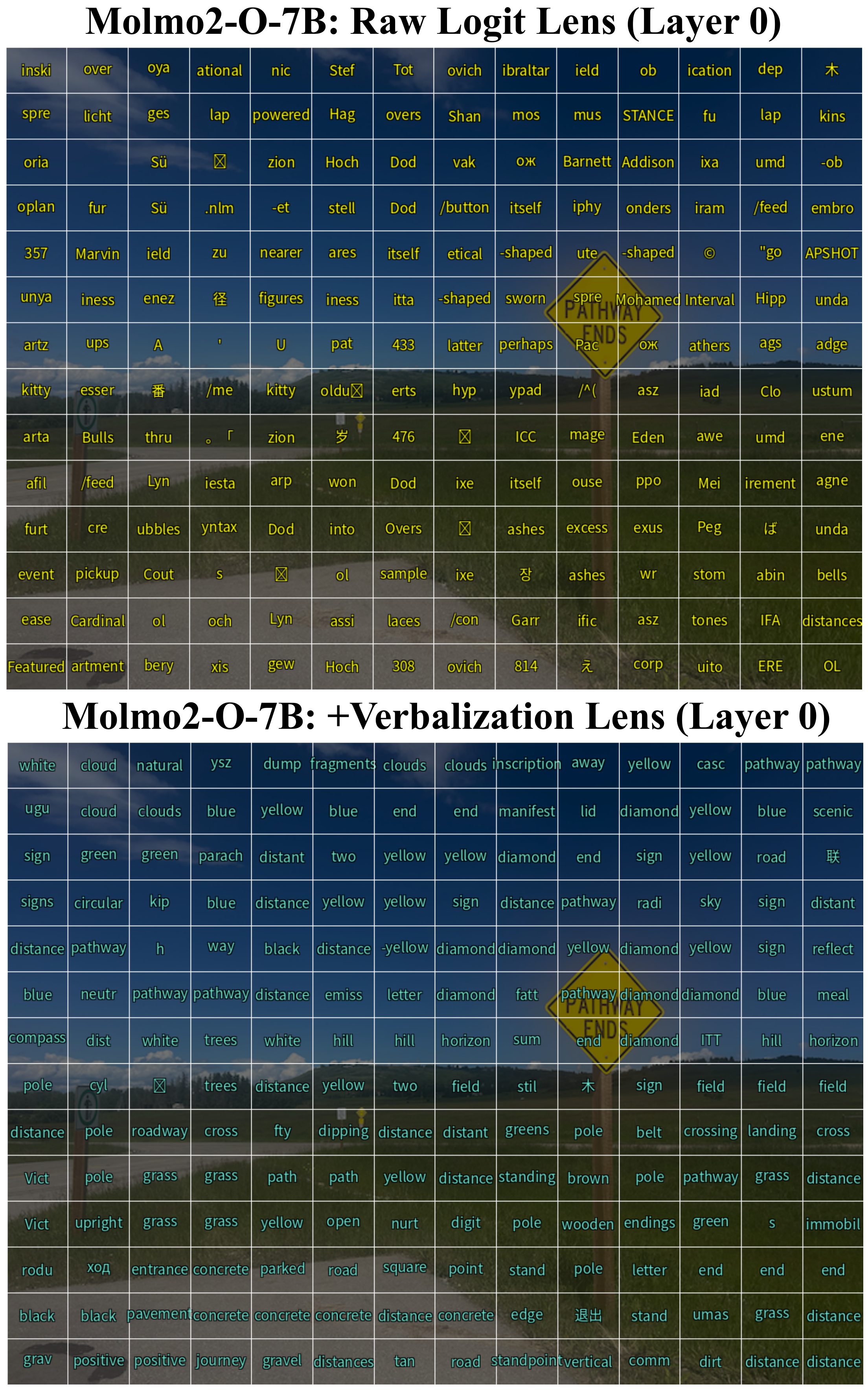}
    \caption{Verbalization lens results for image in Figure~\ref{fig:lens_qual} for \textbf{Molmo2-O-7B}.}
    \label{fig:pathway_molmo2}
\end{figure*}

\begin{figure*}[h]
    \centering
    \includegraphics[width=0.85\linewidth]{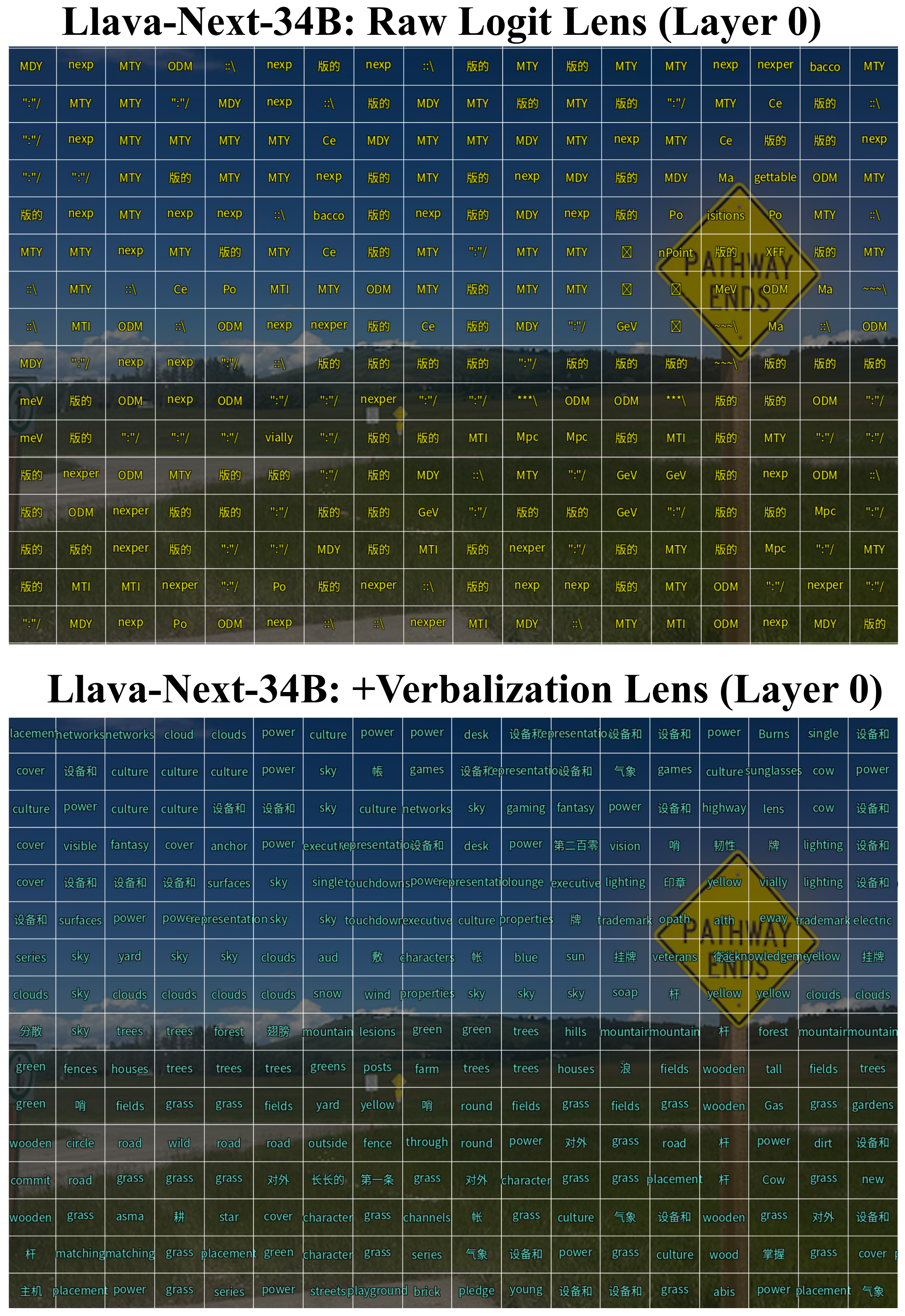}
    \caption{Verbalization lens results for image in Figure~\ref{fig:lens_qual} for \textbf{Llava-Next-34B}. Image is slightly cropped for readability.}
    \label{fig:pathway_llava}
\end{figure*}

\begin{figure*}
    \centering
    \includegraphics[width=\linewidth]{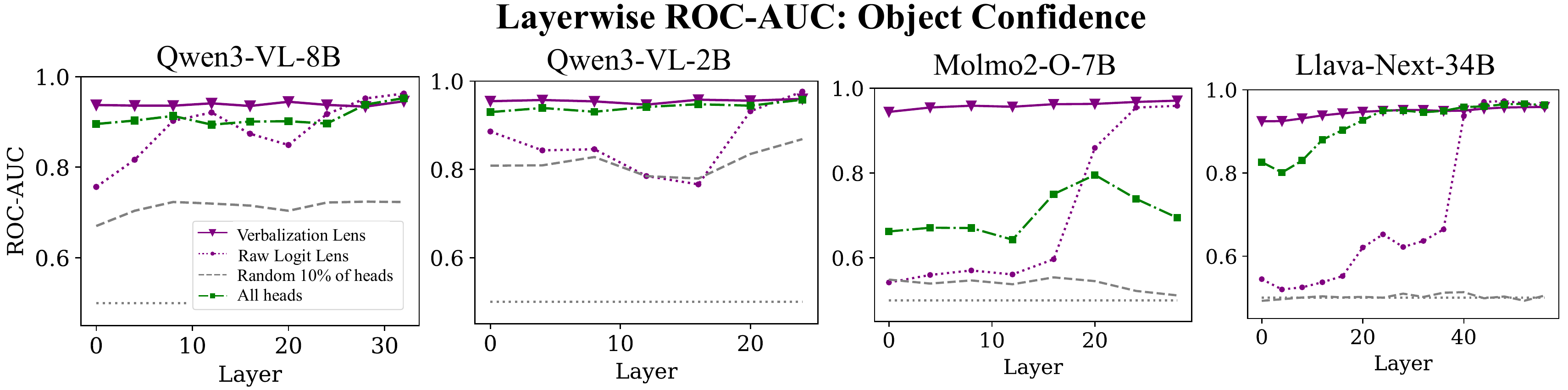}
    \caption{ROC-AUC scores for the experiment in Figure~\ref{fig:detection}b. For raw logit lens, layers where ROC-AUC is high but probability differences in Figure~\ref{fig:detection}b are \emph{low} indicate that logit lens is badly calibrated---even if the ranking between tokens is correct, probability differences are very small, indicating low confidence. We see a similar effect when using all attention heads (a superset that includes our verbalization heads): using all attention heads can achieve high ROC-AUC, but the lower overall probability differences in Figure~\ref{fig:detection}b indicate lower confidence. 
    For verbalization lens, ROC-AUC is close to ceiling across layers in addition to large probability deltas in Figure~\ref{fig:detection}b. This indicates that verbalization lens has higher confidence, which also makes our approach more qualitatively useful (i.e., semantic predictions are assigned high probabilities, making them visible in qualitative visualizations; see interactive demo).}
    \label{fig:roc}
\end{figure*}

\begin{figure*}[h]
    \centering
    \includegraphics[width=\linewidth]{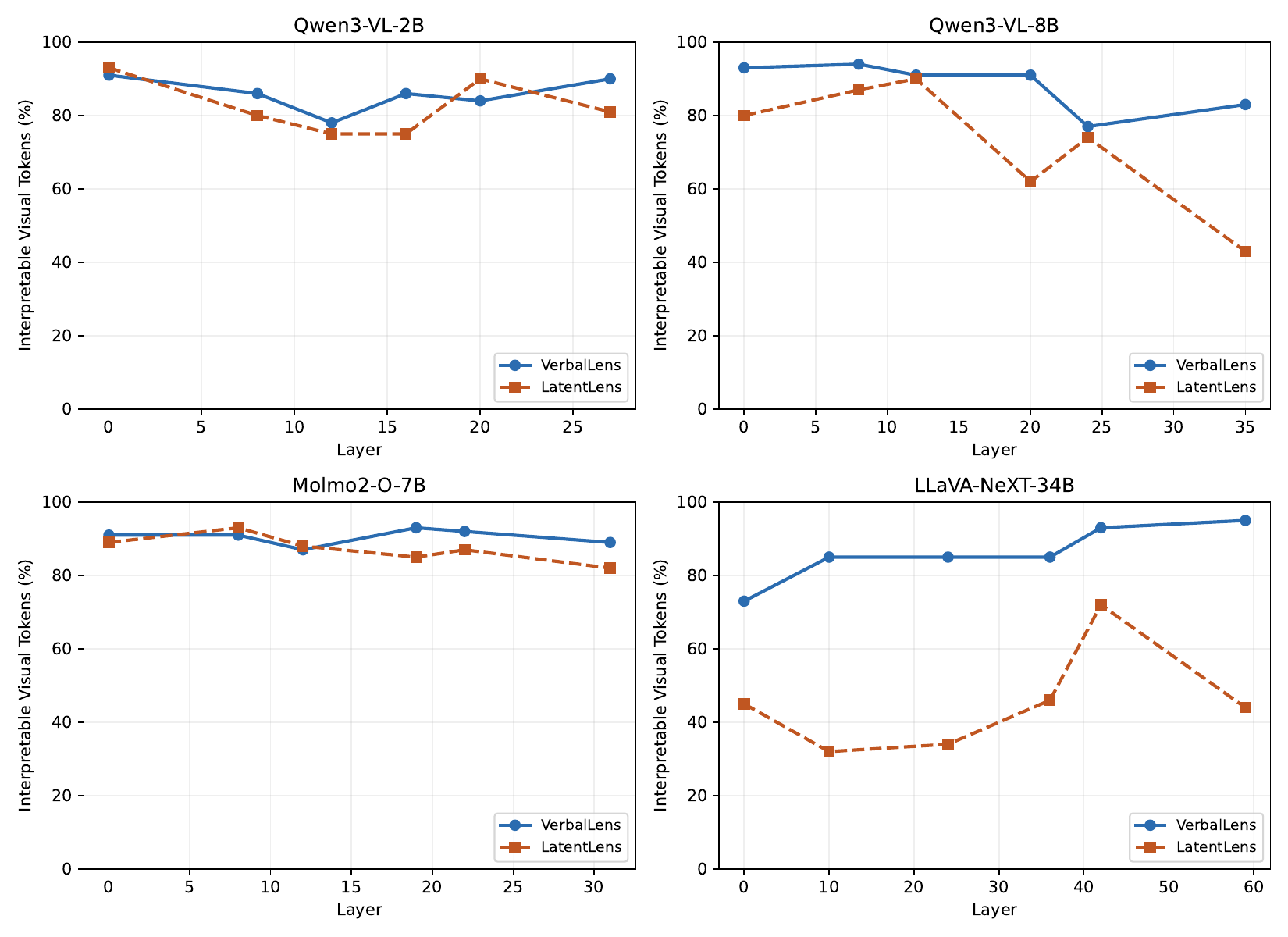}
    \caption{Verbalization lens is on par with or better than LatentLens \cite{latentlens} across all models. We show LLM judge interpretability (\% of patches judged interpretable) across layers (100 images per point). LatentLens shows a pronounced early/mid-layer dip for Llava-Next-34B, but verbalization lens scores consistently high across layers.}
    \label{fig:lens-judge-layers}
\end{figure*}

\begin{figure*}
    \centering
    \includegraphics[width=\linewidth]{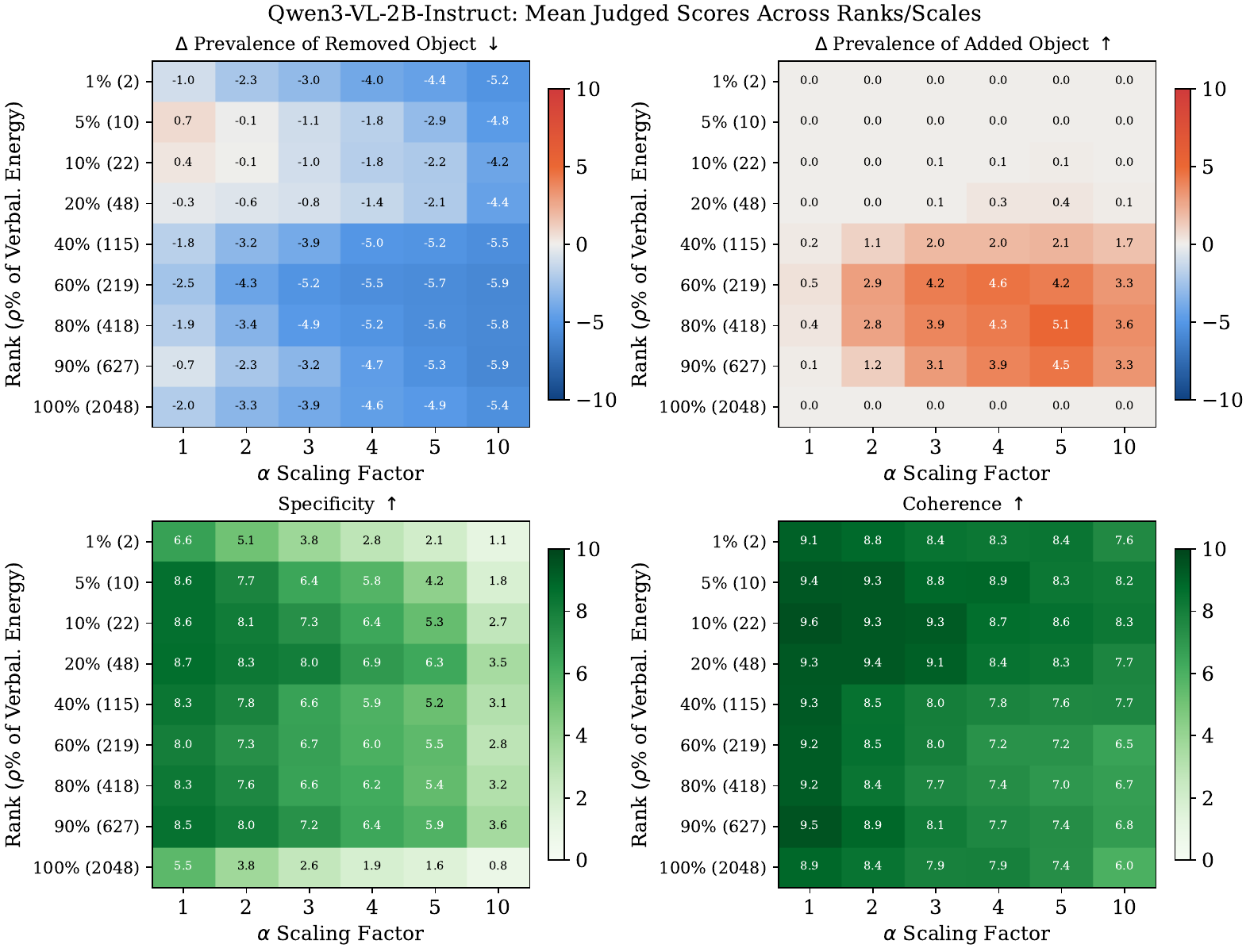}
    \caption{Sweep across pseudo-inverse ranks and scales for editing objects in Qwen3-VL-2B. We generate edits across every combination of rank and scaling factor for 100 random ImageNet images, score the edited captions, judge outputs, and present the mean of scores across images. We focus on results for $\rho$=60\% for Qwen3-VL-2B in the main paper.}
    \label{fig:edit_sweep_2b}
\end{figure*}

\begin{figure*}
    \centering
    \includegraphics[width=\linewidth]{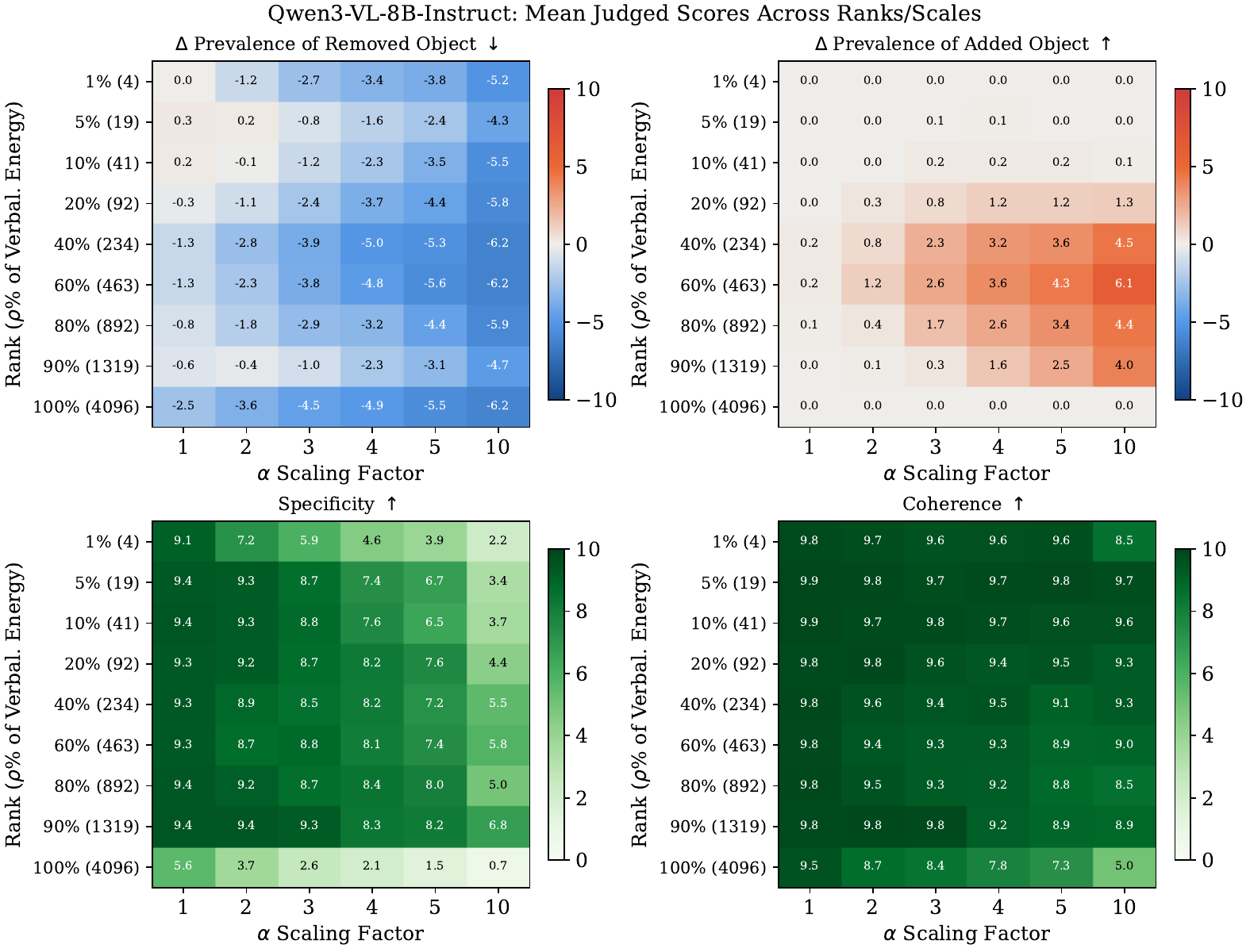}
    \caption{Sweep across pseudo-inverse ranks and scales for editing objects in Qwen3-VL-8B. We generate edits across every combination of rank and scaling factor for 100 random ImageNet images, score the edited captions, judge outputs, and present the mean of scores across images. See Section~\ref{sec:edit_results} for details. Qwen3-VL-8B appears to require a larger scaling factor $\alpha$. We focus on results for rank $\rho=$60\% for Qwen3-VL-8B in the main paper.}
    \label{fig:edit_sweep_8b}
\end{figure*}

\end{document}